\documentclass[10pt,a4paper]{article}
\newcommand{\BrandPrimaryHex}{2452D6}
\newcommand{\BrandTintHex}{F1F4F8}
\newcommand{\BrandTextHex}{000000}
\newcommand{\BrandMutedHex}{000000}
\newcommand{\BrandPanelHex}{2452D6}
\newcommand{\BrandName}{Symbiosis Robotics}
\newcommand{\BrandLogoPath}{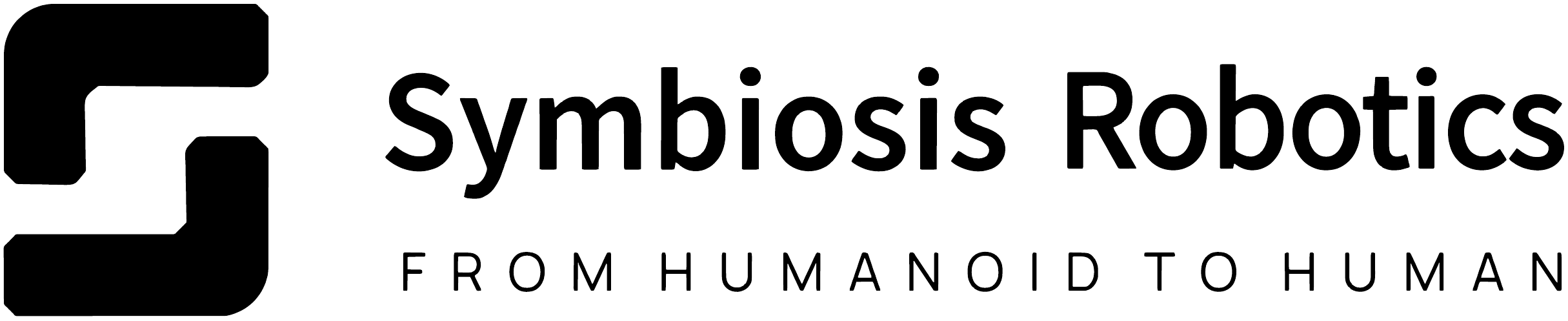}
\newcommand{\BrandLogoWidth}{42mm}
\newcommand{\BrandLogoHeight}{8.6mm}
\newcommand{\BrandMarkPath}{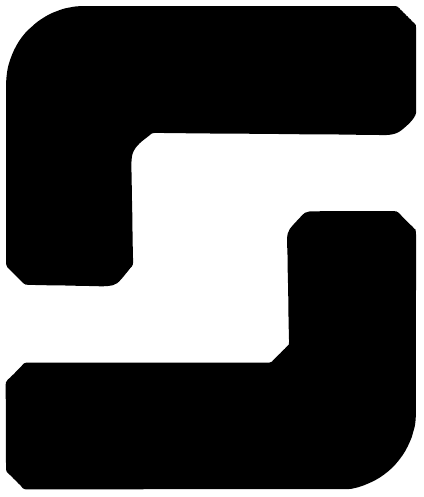}
\newcommand{\BrandMarkHeight}{4mm}

\DeclareRobustCommand{\reportchange}[1]{\textcolor{black}{#1}}

\usepackage{report-template}
\newcommand{\ReportTitle}{HUMANO: A Large-Scale Benchmark for Generalist Humanoid Loco-Manipulation}
\newcommand{\ReportDisplayTitle}{\reportchange{\ReportTitle}}
\newcommand{\ReportSubtitle}{}

\newcommand{\ReportAuthorNames}{Jiajun Fu, Shuanghao Bai, Youyi Kou, Changjing He, Baohua Yin, Hengtao Li, Jiacheng Liu, Peiyang Li, Yulong Wu, Zijian Yang, Qishen Wang, Mingyang Sun, Junyu Wu, Yao Mu, Fuchun Sun, Badong Chen, Pengxiang Ding}
\newcommand{\ReportAuthors}{%
  \textbf{\reportchange{Core Contributors}}\par
  \mbox{\reportchange{Jiajun Fu}}\quad
  \mbox{\reportchange{Shuanghao Bai}}\quad
  \mbox{\reportchange{Youyi Kou}}\quad
  \mbox{\reportchange{Changjing He}}\par\smallskip

  \textbf{\reportchange{Project Lead}}\par
  \mbox{\reportchange{Shuanghao Bai}}\quad
  \mbox{\reportchange{Jiajun Fu}}\par\smallskip
  
  \textbf{\reportchange{Contributors}}\par
  \mbox{\reportchange{Baohua Yin}}\quad
  \mbox{\reportchange{Hengtao Li}}\quad
  \mbox{\reportchange{Jiacheng Liu}}\quad
  \mbox{\reportchange{Peiyang Li}}\quad
  \mbox{\reportchange{Yulong Wu}}\quad
  \mbox{\reportchange{Zijian Yang}}\quad
  \mbox{\reportchange{Qishen Wang}}\quad
  \mbox{\reportchange{Mingyang Sun}}\quad
  \mbox{\reportchange{Junyu Wu}}\quad
  \mbox{\reportchange{Yao Mu}}\par\smallskip
  
  \textbf{\reportchange{Advisors}}\par
  \mbox{\reportchange{Badong Chen}}\quad
  \mbox{\reportchange{Pengxiang Ding}}}
\newcommand{\ReportType}{Technical Report}
\newcommand{\ReportVersion}{Template v0.7}

\newcommand{\ReportContact}{}

\usepackage{makecell}
\newcommand{\newer}[1]{\textcolor{black}{#1}} % rebuttal  

\usepackage{fontawesome5}
\usepackage{tikz}
\usepackage{hyperref}

\begin{document}

\begin{reportfrontmatter}
\makereporttitle

% \par\vspace{-0.3em}
% \noindent
% {\sffamily\bfseries
% \hyperref[sec:authors]{\textcolor{black}{Symbiosis Robotics Team}}\par}
% \par\vspace{0.8em}

% % \par\vspace{0.3em}
% \noindent
% {\sffamily\small
% \href{https://symbiosis-robotics.github.io/HUMANO/}{%
%     \reportchange{\faGlobe\ Project Page}}
% \hspace{1.5em}
% \href{https://github.com/Symbiosis-Robotics/HUMANO}{%
%     \reportchange{\faGithub\ Code}}
% \hspace{1.5em}
% \href{https://huggingface.co/datasets/Symbiosis-Robotics/HUMANO}{%
%     \reportchange{\faDatabase\ Dataset}}
% }

\par\vspace{-0.4em}
\noindent
\begin{minipage}[c]{0.42\textwidth}
    {\sffamily\bfseries
    \hyperref[sec:authors]{\textcolor{black}{Symbiosis Robotics Team}}}
\end{minipage}%
\hfill
\begin{minipage}[c]{0.56\textwidth}
    \raggedleft
    {\sffamily\small
    \href{https://symbiosis-robotics.github.io/HUMANO/}{%
        \reportchange{\faGlobe\ Project Page}}
    \hspace{1em}
    \href{https://github.com/Symbiosis-Robotics/HUMANO}{%
        \reportchange{\faGithub\ Code}}
    \hspace{1em}
    \href{https://huggingface.co/datasets/Symbiosis-Robotics/HUMANO}{%
        \reportchange{\faDatabase\ Dataset}}
    }
\end{minipage}
\par\vspace{3mm}

\begin{reportabstract}
Recent advances in vision-language-action models and whole-body control have rapidly expanded the capabilities of humanoid loco-manipulation, enabling generalist policies to perform increasingly diverse tasks with a single model. However, existing benchmarks often focus on individual tasks or small task suites, making it difficult to systematically evaluate multi-task capabilities and generalization. To fill this gap, we present \textsc{HUMANO}, a large-scale benchmark for generalist humanoid loco-manipulation. \textsc{HUMANO} contains 100 tasks organized into eight complementary suites, spanning 90 interactive scenes and over 2K object assets, with 5K in-distribution whole-body demonstrations. The benchmark covers fundamental skills, goal-directed loco-manipulation, generalization across spatial configurations, objects, robot configurations, and goals, as well as long-horizon skill composition and state-dependent reasoning. \textsc{HUMANO} further provides 30K OOD demonstrations covering six non-linguistic distribution shifts (distractors, lighting, background, camera viewpoint, sensor noise, and robot initialization), along with 17.5K evaluation episodes across all seven OOD settings, including language variation.
All policies operate through a shared SONIC-based motion-latent interface, enabling standardized comparison under a common whole-body execution framework. Experiments with representative imitation and vision-language-action policies demonstrate the emerging potential of generalist models for multi-task humanoid loco-manipulation. Although multi-task learning remains more challenging than single-task training, performance continues to improve with increasing demonstration scale. Preliminary sim-to-real results show that policies trained on HUMANO simulation data can transfer executable whole-body behaviors to a physical humanoid robot.
\end{reportabstract}
\end{reportfrontmatter}

\begin{figure*}[htbp]
    \centering
    \includegraphics[width=\textwidth]{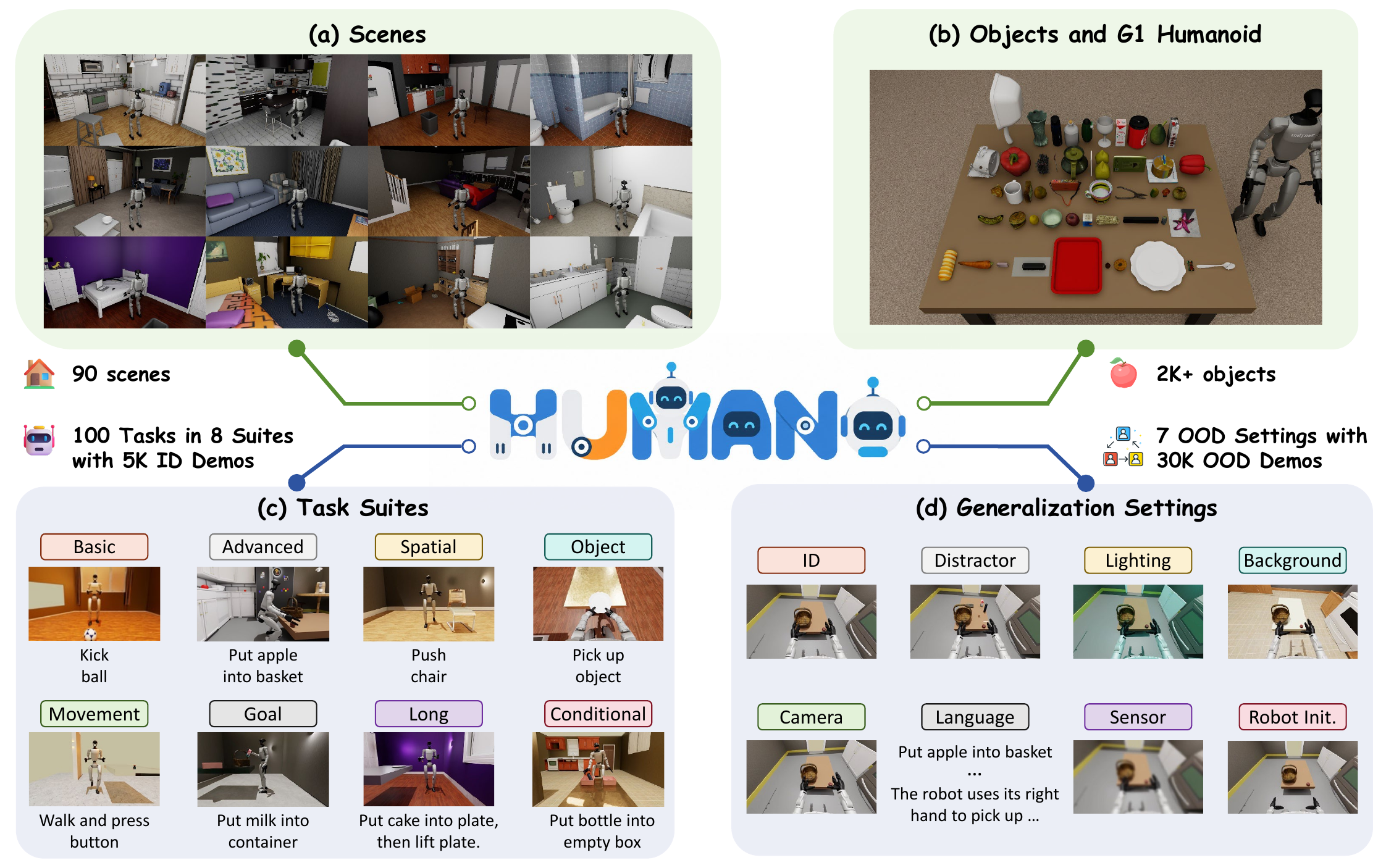}
    \vskip -0.1in
    \caption{
    Overview of \textsc{HUMANO}. The benchmark contains 100 tasks across 8 capability suites, spanning 90 interactive scenes and over 2K object assets, together with 5K in-distribution demonstrations. \reportchange{Six OOD conditions provide an additional 30K demonstrations, bringing the total to 35K. HUMANO-Plus evaluates policies under seven controlled distribution shifts.}
    }
    \label{fig:teaser}
\end{figure*}

\section{Introduction}
\label{sec:introduction}

Recent advances in vision-language-action models and whole-body control have substantially accelerated progress in humanoid loco-manipulation, enabling increasingly generalist policies to integrate perception, language understanding, locomotion, and manipulation within a single model~\citep{bjorck2025gr00t,intelligence2025pi05,wei2026psi0,luo2026sonic}. This progress is moving humanoid robots beyond task-specific behaviors toward general-purpose systems capable of performing diverse whole-body tasks in complex environments. However, the evaluation of such systems still relies heavily on a relatively small number of real-robot tasks. While real-world evaluation is indispensable, it is costly, time-consuming, difficult to reproduce, and inherently limited in task and environment coverage. Simulation provides a complementary route for scalable and standardized evaluation, allowing policies to be tested efficiently across substantially larger task sets, diverse environments, and controlled distribution shifts. As humanoid policies become increasingly general, building such comprehensive simulation-based evaluation has therefore become an important prerequisite for measuring their true capabilities and generalization.

Existing humanoid simulation benchmarks have begun to address this need, but each captures only a subset of the capabilities required by generalist loco-manipulation policies. Some benchmarks primarily emphasize reinforcement learning (RL) and whole-body control~\citep{sferrazza2024humanoidbench}, while others expand task and scene diversity for imitation learning and generalist policies~\citep{wei2026simple}. Recent efforts further study specialized aspects such as scalable demonstration generation~\citep{lin2026humanoidmimicgen} or coordination between high-level policies and low-level RL-based whole-body controllers~\citep{wang2026humanoidarena}. Nevertheless, existing benchmarks remain limited along one or more key dimensions, including task scale, capability diversity, systematic generalization, and standardized policy interfaces. As summarized in Table~\ref{tab:benchmark_comparison}, a unified benchmark that brings these dimensions together is still missing, making it difficult to comprehensively compare the rapidly growing class of generalist humanoid loco-manipulation policies.

To bridge this gap, we introduce \textsc{HUMANO}, a large-scale benchmark for evaluating generalist humanoid loco-manipulation, as illustrated in Figure~\ref{fig:teaser}. \textsc{HUMANO} contains 100 tasks organized into eight complementary suites, spanning 90 interactive scenes and more than 2K object assets, together with 5K in-distribution whole-body demonstrations. The benchmark covers a broad spectrum of humanoid capabilities: \emph{Basic} and \emph{Advanced} evaluate fundamental whole-body skills and goal-directed loco-manipulation; \emph{Spatial}, \emph{Object}, \emph{Movement}, and \emph{Goal} probe generalization across spatial configurations, object variations, movement requirements, and task goals; and \emph{Long-Horizon} and \emph{Conditional} evaluate temporal skill composition and state-dependent decision making. Rather than increasing task count through minor variations, this structured design organizes distinct forms of whole-body behavior into a unified benchmark, enabling systematic evaluation of a single policy across diverse humanoid tasks.

Beyond task diversity, \textsc{HUMANO} introduces \textsc{HUMANO-Plus}, an OOD robustness benchmark comprising seven controlled distribution shifts: distractors, lighting, scene appearance, camera viewpoint, language variation, visual sensor corruption, and robot initialization. \reportchange{Each shift has 25 evaluation episodes per task, for a total of 17.5K OOD episodes. Policies train on the in-distribution demonstrations.}
To further standardize policy comparison, \textsc{HUMANO} adopts the unified motion representation of SONIC~\citep{luo2026sonic} as the interface between high-level policy prediction and whole-body execution. Representative imitation policies and vision-language-action models therefore operate through the same motion-latent interface and share a common whole-body decoder, reducing confounding from heterogeneous low-level control stacks. We benchmark these models across all 100 tasks and seven OOD settings, and further study demonstration scaling, single-task versus multi-task learning, and sim-to-real transfer.

\begin{table*}[t]
\centering
\small
\caption{
Comparison of representative humanoid loco-manipulation benchmarks in benchmark scale and generalization coverage.
Generalization spans six categories: visual, language, spatial, semantic, sensing, and temporal.
$\checkmark$ indicates explicit benchmark-level coverage, while -- indicates that the corresponding aspect is not a primary component.
$^\dagger$ denotes simulated teleoperation episodes.
}
\label{tab:benchmark_comparison}

\vskip -0.05in

\resizebox{\textwidth}{!}{
\begin{tabular}{lcccccccccc}
\toprule
& & & & &
\multicolumn{6}{c}{Generalization} \\
\cmidrule(lr){6-11}
Benchmark &
\# Tasks &
\# Scenes &
\# Objects &
\# Demos &
Visual &
Lang. &
Spatial &
Semantic &
Sensing &
Temporal \\
\midrule

HumanoidBench~\cite{sferrazza2024humanoidbench}
& 27
& --
& 20+
& --
& --
& --
& --
& --
& --
& -- \\

SIMPLE~\cite{wei2026simple}
& 60
& 50
& 1K+
& 6K
& \checkmark
& \checkmark
& \checkmark
& \checkmark
& --
& -- \\

HumanoidMimicGen~\cite{lin2026humanoidmimicgen}
& 9
& --
& --
& 9K
& --
& --
& \checkmark
& \checkmark
& --
& -- \\

HumanoidArena~\cite{wang2026humanoidarena}
& 7
& --
& --
& 1.4K
& \checkmark
& --
& --
& \checkmark
& --
& \checkmark \\

LocoManiLab~\cite{wang2026locomanilab}
& 234
& 13
& --
& 14K+$^\dagger$
& \checkmark
& --
& \checkmark
& \checkmark
& --
& \checkmark \\

\midrule

\textbf{\textsc{HUMANO} (Ours)}
& \textbf{100}
& \textbf{90}
& \textbf{2K+}
& \textbf{\reportchange{35K}}
& \checkmark
& \checkmark
& \checkmark
& \checkmark
& \checkmark
& \checkmark \\

\bottomrule
\end{tabular}
}
\end{table*}

In summary, \textsc{HUMANO} provides a large-scale benchmark for generalist humanoid loco-manipulation, a systematic protocol for evaluating both multi-task competence and OOD generalization, and a unified motion-latent interface for controlled comparison of high-level policies. Through comprehensive evaluation and analysis, \textsc{HUMANO} establishes a standardized testbed for measuring progress toward more general and robust humanoid loco-manipulation.

\section{Related Work}

\subsection{Humanoid Loco-Manipulation Models}

Modern humanoid loco-manipulation systems commonly adopt hierarchical architectures that separate high-level policy prediction from low-level whole-body execution. At the high level, imitation-learning methods such as ACT~\citep{zhao2023learning} and Diffusion Policy~\citep{chi2025diffusion}, together with generalist foundation policies such as $\pi_{0.5}$~\citep{intelligence2025pi05}, UnifoLM-WLA-1.0~\citep{unitree2026unifolmwla}, and GR00T N1~\citep{bjorck2025gr00t}, provide general action-prediction capabilities. Recent humanoid-specific models, including WholeBodyVLA~\citep{jiang2026wholebodyvla}, $\Psi_0$~\citep{wei2026psi0}, HEX~\citep{bai2026hex}, and OpenHLM~\citep{hu2026openhlm}, extend VLA learning to coordinated locomotion and manipulation. HAF-VLA~\citep{gu2026haf} further introduces progressive action generation across locomotion, waist, and manipulation stages, while Holo-M~\citep{shao2026holom} explores discrete action representations for whole-body control. In parallel, MotionWAM~\citep{zheng2026motionwam}, $\omega$-0~\citep{li2026omega}, and WholeBodyWAM~\citep{zhang2026wholebodywam} incorporate predictive motion and world dynamics into whole-body policy learning. Another WholeBodyWAM~\citep{li2026wholebodywam} adapts pre-trained world-action priors specifically for humanoids through WBC-grounded coordination.

Beyond policy architectures, recent efforts improve humanoid learning through large-scale visual sim-to-real transfer~\citep{he2025viral} and human motion data~\citep{hu2026workhorse,engineai2026awakenhi}. TrajBooster~\citep{liu2026trajbooster} further leverages cross-embodiment trajectory retargeting to generate humanoid-compatible demonstrations for VLA training. Meanwhile, foundation models such as $\lambda_0$~\citep{lambda2026lambda0}, HuGe~\citep{liu2026huge}, and GR00T 2 Turbo~\citep{nvidia2026gr00t2} adopt SONIC motion tokens as intermediate representations for whole-body execution. Unlike these predominantly hierarchical approaches, DPC~\citep{symbiosis2026directperceptioncontrol} directly predicts joint positions and hand targets without a separate learned whole-body controller.

At the low level, DeepMimic~\citep{peng2018deepmimic} and AMP~\citep{peng2021amp} establish physics-based motion imitation, while BeyondMimic~\citep{liao2026beyondmimic} scales motion tracking and composition to diverse humanoid behaviors. HOMIE~\citep{ben2025homie}, AMO~\citep{li2025amo}, CLONE~\citep{li2025clone}, and TWIST2~\citep{ze2026twist2} extend whole-body control toward teleoperation and loco-manipulation. More recent approaches, including HLO-AMP~\citep{sleiman2026hloamp} and Helix 02~\citep{figure2026helix02}, further explore interaction-aware motion priors and learned whole-body control. SONIC~\citep{luo2026sonic} provides a reusable motion representation and whole-body decoder for scalable humanoid motion tracking. Building on this interface, HUMANO standardizes whole-body execution across heterogeneous high-level policies, while also recording joint-level and hand commands to support non-hierarchical policy learning.

\subsection{Humanoid Data Generation and Simulation}

Collecting whole-body humanoid interaction data directly on physical robots remains costly and labor-intensive. Early approaches rely primarily on teleoperation, motion capture, and human-to-humanoid retargeting, as demonstrated by HumanPlus~\citep{fu2024humanplus}, OmniH2O~\citep{he2024omnih2o}, and TWIST~\citep{ze2025twist}. To improve scalability, TWIST2~\citep{ze2026twist2} introduces portable, mocap-free teleoperation, while HuMI~\citep{nai2026humi} enables humanoid-free demonstrations. VideoMimic~\citep{allshire2025videomimic} and EgoLoAF~\citep{zeng2026egoloaf} further explore learning from human videos and egocentric demonstrations. In parallel, simulation-based approaches scale data generation through automated motion synthesis and adaptation. HumanoidMimicGen~\citep{lin2026humanoidmimicgen} generates demonstrations through whole-body planning and adaptation, while InterMimicGen~\citep{zhang2026intermimicgen} employs self-evolving motion imitation. MotionDisco~\citep{taouil2026motiondisco} discovers contact-rich motions through LLM-guided evolutionary search and kinodynamic optimization. Complementary video-based approaches, including PRISM~\citep{wang2026counterfactual} and GRAIL~\citep{xie2026grail}, leverage counterfactual video synthesis and video world models to generate diverse human-object interactions and robot-compatible motions.

Simulation platforms further seek to bridge physical and visual fidelity for humanoid policy learning. SIMPLE~\citep{wei2026simple} decouples physics simulation from visual rendering, combining contact-rich dynamics in MuJoCo with photorealistic rendering in Isaac Sim. OASIS~\citep{yu2026oasis} integrates asset reconstruction, simulated data collection, and domain randomization for sim-to-real transfer. LEGS~\citep{kim2026legs} leverages 3D Gaussian Splatting for photorealistic rendering and appearance augmentation, while Zero-HLM~\citep{yang2026zerohlm} combines real-to-sim reconstruction with motion-environment co-adaptation to generate physically feasible demonstrations for zero-shot sim-to-real transfer. Following the decoupled simulation and rendering paradigm, HUMANO integrates SONIC-based whole-body control in MuJoCo with high-fidelity offline rendering in Isaac Sim. While these works primarily advance data generation and simulation, HUMANO complements them with standardized evaluation across diverse tasks, controlled distribution shifts, and sim-to-real settings.

\subsection{Humanoid Loco-Manipulation Benchmarks}

General-purpose platforms such as RoboVerse~\citep{geng2025roboverse} and whole-body imitation-learning benchmarks such as LocoMuJoCo~\citep{al2023locomujoco} provide infrastructure for humanoid robot learning. HumanoidBench~\citep{sferrazza2024humanoidbench} introduces 27 whole-body locomotion and manipulation tasks, while SIMPLE~\citep{wei2026simple} provides a unified simulation testbed with progressive domain randomization for policy learning and evaluation. Beyond task-level evaluation, HumanoidMimicGen~\citep{lin2026humanoidmimicgen} focuses on scalable demonstration generation, whereas HumanoidArena~\citep{wang2026humanoidarena} studies perturbation robustness and policy-tracker transfer. Other benchmarks target specific capabilities, including cross-embodiment loco-manipulation in SkillBench~\citep{kuang2025skillblender}, tool selection and mobile execution in HumanoidToolBench~\citep{jang2026humanoidtoolbench}, and real-world policy evaluation through the cloud-based platform of Humanoid Everyday~\citep{zhao2025humanoideveryday}. More recently, LocoManiLab~\citep{wang2026locomanilab} integrates data generation, policy learning, evaluation, and physical deployment into a unified testbed, covering atomic and compositional tasks, unseen tasks and scenes, controller variations, and sim-to-real transfer.

HUMANO complements these efforts with a capability-oriented benchmark comprising 100 tasks across eight suites, 90 interactive scenes, and over 2K object assets. Its task organization systematically evaluates fundamental whole-body skills, generalization across spatial configurations, objects, robot configurations, and goals, as well as long-horizon skill composition and state-dependent decision-making. Furthermore, HUMANO-Plus introduces seven individually controlled OOD shifts across all eight suites, enabling systematic assessment of robustness to visual, linguistic, sensory, and robot-initialization variations. Beyond evaluation, HUMANO provides 30K OOD demonstrations spanning five visual/sensor conditions and one language condition, in addition to 5K ID demonstrations, yielding 35K demonstrations in total. Together, these designs enable systematic assessment of multi-task competence, capability generalization, and robustness under controlled distribution shifts.

% 家俊 & 昌璟

\section{The HUMANO Benchmark}

\textsc{HUMANO} contains 100 tasks across eight capability suites, 90 interactive scenes, over 2K object assets, and \reportchange{35K demonstrations. These include 5K in-distribution demonstrations and 30K OOD demonstrations derived from the same trajectories across five visual conditions and one language condition}. It further provides seven controlled OOD settings and a unified SONIC-based motion-latent interface. Section~\ref{subsec:assets} introduces the environment assets, Section~\ref{subsec:tasks} describes the task-suite design, Section~\ref{subsec:generalization} presents the generalization settings, Section~\ref{subsec:collection} details whole-body demonstration collection and the unified policy interface, and Section~\ref{subsec:dataset} describes dataset construction and statistics.

\subsection{Environment Assets}
\label{subsec:assets}

HUMANO builds its environments on the iTHOR scene collection distributed through MolmoSpaces~\citep{kim2026molmospaces}. iTHOR contains 120 hand-crafted interactive indoor scenes, evenly divided among kitchens, living rooms, bedrooms, and bathrooms. Since HUMANO does not target bathroom-specific activities, we exclude the 30 bathroom scenes and retain the remaining 90 scenes, comprising 30 kitchens, 30 living rooms, and 30 bedrooms. These scenes provide diverse indoor layouts, furniture arrangements, and interactive structures for constructing humanoid loco-manipulation tasks.

HUMANO further draws object assets from Synthesis, MolmoSpaces, and Objaverse, yielding more than 2K indoor objects with diverse geometries and semantic categories. The asset pool includes both manipulable objects and scene-level furniture, supporting tasks involving grasping, placement, rearrangement, articulated interaction, and whole-body movement in cluttered indoor environments. Together, the 90 scenes and 2K+ object assets provide the environment basis for the task suites and controlled generalization settings introduced below.

\subsection{Task Suite Design}
\label{subsec:tasks}

\begin{figure}[t]
\centering
\includegraphics[width=\linewidth]{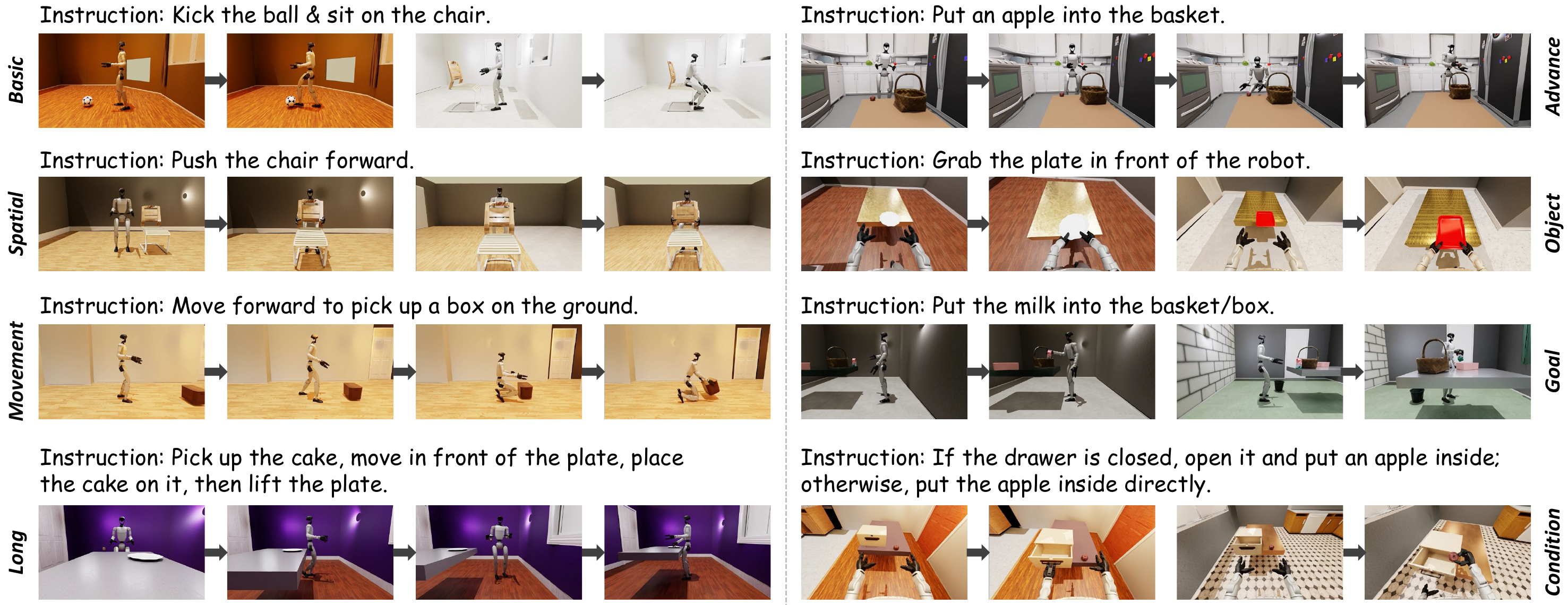}
\vskip -0.05in
\caption{
Representative tasks from the 8 HUMANO suites. Each row shows an example instruction and keyframes of the corresponding whole-body behavior, covering fundamental skills, loco-manipulation, generalization, long-horizon composition, and conditional reasoning.
}
\label{fig:suite}
\end{figure}

As shown in Figure~\ref{fig:suite}, HUMANO builds on the environment assets described above to construct 100 humanoid loco-manipulation tasks organized into eight complementary suites. Rather than relying on minor task variations, the suites are structured around distinct capability dimensions, spanning fundamental skills, goal-directed manipulation, generalization, temporal composition, and conditional decision making. Complete definitions of all 100 tasks are provided in Appendix~\ref{app:tasks}.

\textbf{Basic Suite.}
The \emph{Basic} suite contains 10 tasks and evaluates fundamental whole-body skills, including grasping, pushing, posture changes, reaching, and lower-body interaction.

\textbf{Advanced Suite.}
The \emph{Advanced} suite contains 10 tasks and extends these primitives to goal-directed loco-manipulation involving object transport, placement, and coordinated body-object interaction.

\textbf{Spatial Suite.}
The \emph{Spatial} suite contains 20 tasks and evaluates generalization to variations in target layout and interaction height relative to the robot.

\textbf{Object Suite.}
The \emph{Object} suite contains 10 tasks and evaluates whether manipulation skills transfer across different object categories and geometries while preserving the underlying task semantics.

\textbf{Movement Suite.}
The \emph{Movement} suite contains 10 tasks and evaluates robustness to variations in whole-body movement requirements while preserving the underlying manipulation objective.

\textbf{Goal Suite.}
The \emph{Goal} suite contains 10 tasks and evaluates goal-conditioned generalization by varying the desired target or destination while maintaining comparable task contexts.

\textbf{Long-Horizon Suite.}
The \emph{Long-Horizon} suite contains 20 tasks and evaluates temporal composition by chaining multiple locomotion and manipulation primitives into extended task sequences.

\textbf{Conditional Suite.}
The \emph{Conditional} suite contains 10 tasks and evaluates state-dependent decision making, where the required action sequence changes according to the observed environment state.

\subsection{Generalization Settings}
\label{subsec:generalization}

As discussed in Section~\ref{subsec:tasks}, all HUMANO task suites already incorporate generalization challenges through variations in object spatial configuration, object identity, task goals, and temporal composition. To further assess robustness beyond these suite-specific variations, we introduce \textsc{HUMANO-Plus}, an extended OOD benchmark featuring seven controlled distribution shifts consistently applied across all eight task suites: \emph{Distractor}, \emph{Lighting}, \emph{Background}, \emph{Camera Viewpoint}, \emph{Language}, \emph{Sensor Noise}, and \emph{Robot Initial State}. These settings systematically evaluate model robustness to changes in visual appearance, linguistic expression, sensing quality, and robot initialization while preserving the underlying task objectives. Detailed definitions and implementation procedures are provided in Appendix~\ref{app:ood_settings}.

\textbf{Distractor.}
We introduce \reportchange{five} task-irrelevant objects near the interaction region while excluding objects that overlap semantically with task-relevant categories, testing robustness to visual and physical clutter.

\textbf{Lighting.}
We vary illumination through changes in light color, spatial distribution, and \reportchange{intensity}, evaluating robustness to diverse lighting conditions.

\textbf{Background.}
We evaluate transfer to unseen scene layouts and appearance configurations, testing whether policies rely on task-relevant cues rather than memorized environments.

\textbf{Camera Viewpoint.}
We perturb the extrinsics of the head-mounted cameras through \reportchange{joint yaw and pitch rotations}, measuring robustness to observation-view changes.

\textbf{Language.}
We replace the original instruction with semantically equivalent expressions that vary wording and contextual description while preserving the intended task.

\textbf{Sensor Noise.}
We perturb RGB observations using controlled Gaussian blur with \reportchange{a standard deviation in the range of 1--10 pixels} to evaluate robustness to degraded visual sensing.

\textbf{Robot Initial State.}
We perturb the initial base position and orientation within a valid local region while rejecting configurations that introduce collisions or invalid task states.

Under the standard HUMANO protocol, the seven OOD settings are used only for evaluation: models are trained on in-distribution demonstrations and directly tested under each controlled shift. \reportchange{The expansion yields 30K} OOD demonstrations covering \reportchange{six} settings for optional adaptation and robustness studies. \reportchange{The Language condition pairs each original trajectory with a rephrased instruction.} Robot Initial State is excluded because corresponding demonstrations cannot be generated through the same rendering pipeline and related configuration variation is already covered by the Spatial suite.

\subsection{Demonstration Collection} 
\label{subsec:collection}

\begin{figure*}[t]
\centering
\includegraphics[width=\textwidth]{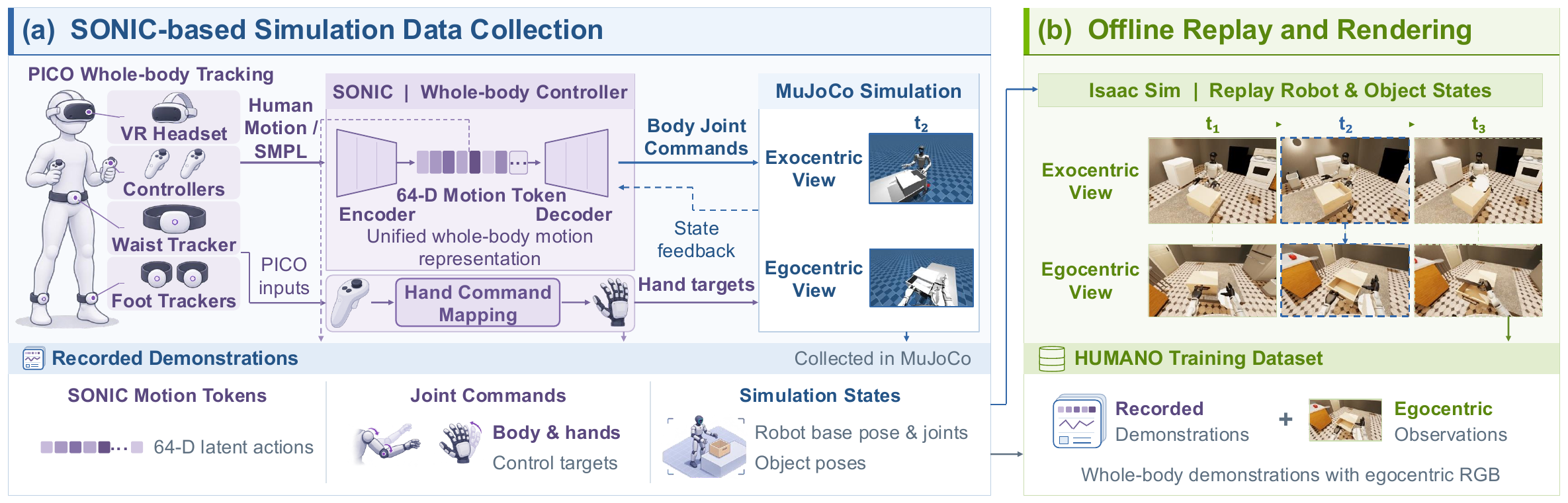}
\vskip -0.03in
\caption{
Overview of the HUMANO \reportchange{SONIC-based simulation} data-collection and rendering pipeline.
(a) Human motion captured with a PICO VR system is encoded into SONIC motion-latent tokens and decoded into body joint commands for execution in MuJoCo, with hand commands provided separately by the PICO controls. \reportchange{SONIC motion tokens}, \reportchange{joint commands}, and simulation states are recorded.
(b) The saved robot and object states are replayed in Isaac Sim to render egocentric RGB observations. The rendered images and recorded demonstration data together form the training dataset.
}
\label{fig:data_collection}
\end{figure*}

\textbf{Whole-Body Teleoperation.} 
As shown in Figure~\ref{fig:data_collection}, HUMANO collects demonstrations through a PICO-based VR teleoperation system coupled with the SONIC whole-body controller~\citep{luo2026sonic}. During teleoperation, we record \reportchange{SONIC motion tokens}, \reportchange{joint commands}, and simulation states. The simulation states capture the robot's base pose and joint configurations, together with the states of target objects. Unlike teleoperation pipelines that separately specify upper-body manipulation and base locomotion commands, our system represents operator motion within a unified whole-body motion space. The captured motion is encoded into SONIC motion-latent tokens and decoded into joint commands for whole-body execution, enabling coordinated locomotion and manipulation through a single control interface. This design supports complex behaviors such as walking, turning, bending, reaching, carrying, and pushing without manually decomposing them into separate locomotion and manipulation commands.
\newer{Beyond SONIC, HUMANO also supports alternative whole-body controllers, including AMO~\cite{li2025amo} and TWIST2~\cite{ze2026twist2}, along with their corresponding data collection pipelines. Furthermore, the system has been adapted to different dexterous hands, including Inspire and BrainCo, enhancing its compatibility with diverse robotic hardware configurations.}

\textbf{Simulation Replay and Rendering.}
HUMANO separates interactive demonstration collection in MuJoCo from offline visual rendering in Isaac Sim. For rendering, we reconstruct the corresponding scene in Isaac Sim and replay these saved states frame by frame. Scenes and task placements are selected to accommodate the recorded trajectories, rejecting conflicts with fixed structures and selectively hiding conflicting background assets. Egocentric RGB observations are then rendered from the robot-mounted camera and combined with the corresponding recorded demonstration data to construct the training dataset. This workflow preserves the \reportchange{selected} motion sequence while allowing visual observations to be generated independently of the teleoperation process.

\textbf{Unified Policy Interface.} 
A central design choice of HUMANO is to use a common SONIC-based whole-body representation throughout demonstration collection, policy learning, and benchmark evaluation. Heterogeneous high-level policies, including imitation-learning methods and vision-language-action models, are adapted to predict \reportchange{SONIC motion tokens together with left- and right-hand commands}. At execution time, \reportchange{the motion tokens} are decoded by a shared SONIC whole-body controller into executable robot motions. \reportchange{The hand commands are applied separately to the corresponding hand joints.} This unified interface provides a consistent action representation and execution pipeline across model families, facilitating policy adaptation and standardized evaluation.

\subsection{Dataset}
\label{subsec:dataset}

\textbf{Dataset Construction.}
Using the above teleoperation system, we collect \reportchange{5K} in-distribution demonstrations covering all 100 tasks across the eight HUMANO suites. %\reportchange{During data curation, we re-execute selected demonstrations in MuJoCo using recorded SONIC motion tokens and hand commands, retaining successful trajectories that remain upright.}
We also provide \reportchange{30K} demonstrations under \reportchange{six of the seven} controlled OOD settings introduced in Section~\ref{subsec:generalization}. These OOD trajectories are kept separate from the default training set and are intended for optional adaptation, robustness, and data-scaling studies. Under the standard HUMANO protocol, policies are trained only on the in-distribution demonstrations and evaluated directly under the corresponding distribution shifts.

\textbf{Dataset Format.}
Each demonstration is stored as a multimodal trajectory containing egocentric RGB observations, language instructions, robot proprioceptive states, \reportchange{SONIC motion tokens, and hand commands}. \reportchange{The action targets concatenate 64-dimensional motion tokens with seven commands per hand, yielding a 78-dimensional vector. Low-level joint commands and simulation states are retained for trajectory inspection and rendering.} We additionally record task and environment metadata, including the task identifier, suite, scene, object configuration, robot initialization, and generalization setting when applicable. This unified representation allows different policy architectures to consume the same observations and predict actions under a common whole-body control interface.

\begin{figure*}[t]
\centering
\includegraphics[width=\textwidth]{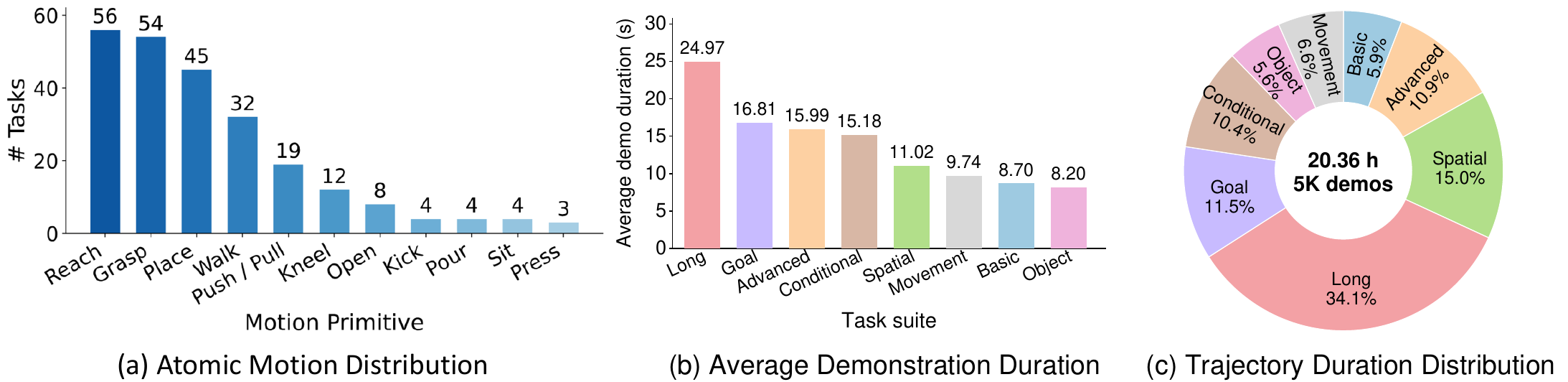}
\vskip -0.05in
\caption{
Dataset statistics of \textsc{HUMANO}. 
(a) Coverage of representative atomic motion primitives across the 100 tasks. 
(b) Average demonstration duration for each task suite. 
(c) Distribution of total trajectory duration across the eight task suites.
}
\label{fig:meta}
\end{figure*}

\textbf{Dataset Statistics.}
As shown in Figure~\ref{fig:meta}, we characterize the HUMANO dataset in terms of motion-primitive coverage and trajectory duration. Figure~\ref{fig:meta}(a) reports the number of tasks involving representative whole-body motion primitives, including reaching, grasping, placing, walking, pushing/pulling, kneeling, opening, kicking, pouring, sitting, and pressing, demonstrating broad coverage of fundamental humanoid behaviors. Figure~\ref{fig:meta}(b) shows the average demonstration duration across the eight task suites, with longer durations generally corresponding to tasks requiring more extended behavioral sequences. Figure~\ref{fig:meta}(c) further summarizes the distribution of total trajectory duration across suites. \reportchange{The 5K demonstrations span 20.36 hours. The Long suite accounts for 34.1\% of the total duration, the largest share of any suite.}

% 小寇

\section{Experiments}

\subsection{Experimental Setup}

\textbf{Baselines.}
We evaluate six representative policy architectures spanning imitation learning and recent generalist robot policies: Diffusion Policy~\citep{chi2025diffusion}, Flow Matching~\citep{bostondynamics2025largebehaviormodelsatlas}, ACT~\citep{zhao2023learning}, $\pi_{0.5}$~\citep{intelligence2025pi05}, GR00T N1.7~\citep{bjorck2025gr00t}, and $\Psi_0$~\citep{wei2026psi0}. All baselines are adapted to the unified SONIC-based motion-latent interface described in Section~\ref{subsec:collection}. Detailed model configurations and training hyperparameters are provided in the appendix.

Detailed model configurations and training settings are provided in Appendix~\ref{app:baseline_implementation}, with the shared evaluation protocol described in Appendix~\ref{app:baseline_evaluation}.

\subsection{Main Results}

Table~\ref{tab:main_results} reports the main in-distribution results on \reportchange{eight} HUMANO suites, covering \reportchange{100} tasks in total. Each baseline is trained independently on each suite using the corresponding in-distribution demonstrations and evaluated with 25 rollouts per task. We report the average success rate for each suite and the \reportchange{unweighted average across all eight suites}.

Overall, the evaluated policies exhibit substantial differences in both absolute performance and capability profiles. GR00T N1.7 achieves the highest overall success rate of \reportchange{28.5\%}, followed by $\pi_{0.5}$ at \reportchange{23.9\%}, substantially outperforming the remaining baselines. On the \emph{Basic} suite, \reportchange{GR00T N1.7 achieves the highest success rate of 21.6\%, followed by $\pi_{0.5}$ at 20.4\%}. \reportchange{GR00T N1.7 also performs best on the \emph{Spatial} and \emph{Goal} suites and ties with $\pi_{0.5}$ on \emph{Object}.} These results demonstrate the promising potential of generalist policies for multi-task humanoid loco-manipulation, while also revealing distinct strengths across capability dimensions.

Performance varies considerably across task suites. The \emph{\reportchange{Object}} suite is relatively more tractable for the strongest policies, whereas success rates are substantially lower on suites requiring \reportchange{goal conditioning or long-horizon skill composition}, particularly \emph{Goal} and \emph{\reportchange{Long}}. Moreover, no single policy dominates across all evaluated suites: $\pi_{0.5}$ performs best on \emph{Advanced} \reportchange{and ties with $\Psi_0$ on \emph{Movement}}, while GR00T N1.7 leads on \emph{Spatial}, \emph{\reportchange{Long}}, and \emph{Goal}. This cross-suite variation highlights the importance of evaluating humanoid policies across multiple capability dimensions rather than relying solely on an aggregate success rate.

\begin{table*}[t]
\centering
\small
\caption{
Main in-distribution results on all 8 HUMANO suites (100 tasks). We report success rate (\%) averaged over 25 rollouts per task. \textbf{Bold} and \underline{underline} denote the best and second-best results, respectively.
}
\label{tab:main_results}
\vskip -0.05in
\resizebox{\textwidth}{!}{
\begin{tabular}{lccccccccc}
\toprule
Method &
Basic &
Advanced &
Spatial &
Object &
\newer{Movement} &
Goal &
\newer{Long} &
Conditional &
Avg. \\
\midrule

Diffusion Policy~\cite{chi2025diffusion}
& 2.8 & 0.4 & 4.6 & 12.4
& \newer{11.6} & 3.6 & \newer{0.6} & 0.4
& \newer{4.6} \\

Flow Matching~\cite{bostondynamics2025largebehaviormodelsatlas}
& 2.0 & 1.2 & 9.8 & 26.4
& \newer{20.0} & 4.4 & \newer{3.2} & 6.8
& \newer{9.2} \\

ACT~\cite{zhao2023learning}
& 4.0 & 2.4 & 9.8 & \underline{32.0}
& \newer{29.6} & 3.6 & \newer{1.2} & 7.2
& \newer{11.2} \\

$\pi_{0.5}$~\cite{intelligence2025pi05}
& \underline{20.4} & \textbf{37.6} & \underline{27.0} & \textbf{41.6}
& \newer{\textbf{38.4}} & \underline{8.8} & \newer{\underline{3.8}} & \underline{13.6}
& \newer{\underline{23.9}} \\

GR00T N1.7~\cite{bjorck2025gr00t}
& \textbf{21.6} & \underline{34.0} & \textbf{35.8} & \textbf{41.6}
& \newer{\underline{36.4}} & \textbf{19.2} & \newer{\textbf{14.6}} & \textbf{24.8}
& \newer{\textbf{28.5}} \\

$\Psi_0$~\cite{wei2026psi0}
& 8.8 & 17.6 & 12.6 & 31.6
& \newer{\textbf{38.4}} & 4.8 & \newer{1.0} & 6.4
& \newer{15.2} \\
\bottomrule
\end{tabular}
}
\vskip -0.1in
\end{table*}

\subsection{Out-of-Distribution Generalization Study}

\begin{table*}[t]
\centering
\small
\caption{
OOD generalization on \textsc{HUMANO-Plus}, covering all \newer{8 HUMANO task suites (100 tasks)} across seven controlled OOD settings. Each task has 25 rollouts per setting. \reportchange{Each column gives the mean success rate (\%) across the eight suites, and OOD Avg. averages the seven settings.} \reportchange{Within each column, \textbf{bold} and \underline{underline} mark the two highest distinct positive scores, respectively.}
}
\label{tab:ood_results}
\vskip -0.05in
\resizebox{\textwidth}{!}{
\begin{tabular}{lcccccccc}
\toprule
Method &
Distractor &
Lighting &
Background &
Camera &
Language &
Sensor &
Robot Init. &
OOD Avg. \\
\midrule

Diffusion Policy~\cite{chi2025diffusion}
& \newer{3.9} & \newer{1.7} & \newer{4.8} & \newer{0.2}
& \newer{4.3} & \newer{0.4} & \newer{2.1}
& \newer{2.5} \\

Flow Matching~\cite{bostondynamics2025largebehaviormodelsatlas}
& \reportchange{5.9} & \reportchange{3.6} & \reportchange{9.5} & \reportchange{2.4}
& \reportchange{9.3} & \reportchange{1.6} & \reportchange{7.7}
& \reportchange{5.7} \\

ACT~\cite{zhao2023learning}
& \newer{5.5} & \newer{2.5} & \newer{10.0} & \newer{0.8}
& \newer{11.3} & \newer{0.6} & \newer{7.6}
& \newer{5.4} \\

$\pi_{0.5}$~\cite{intelligence2025pi05}
& \newer{\underline{21.1}} & \newer{\underline{20.1}}
& \newer{\underline{22.0}} & \newer{\textbf{14.3}}
& \newer{\underline{23.0}} & \newer{\underline{8.1}}
& \newer{\underline{15.7}}
& \newer{\underline{17.8}} \\

GR00T N1.7~\cite{bjorck2025gr00t}
& \newer{\textbf{21.8}} & \newer{\textbf{24.7}}
& \newer{\textbf{28.5}} & \newer{\underline{11.0}}
& \newer{\textbf{26.9}} & \newer{\textbf{10.3}}
& \newer{\textbf{17.0}}
& \newer{\textbf{20.0}} \\

$\Psi_0$~\cite{wei2026psi0}
& \newer{10.4} & \reportchange{12.3} & \newer{12.8} & \newer{6.8}
& \newer{13.2} & \newer{2.5} & \reportchange{9.1}
& \newer{9.6} \\

\bottomrule
\end{tabular}
}
\end{table*}

We next evaluate the robustness of all baselines on \textsc{HUMANO-Plus}, an OOD robustness benchmark that extends HUMANO with seven controlled distribution shifts: \emph{Distractor}, \emph{Lighting}, \emph{Background}, \emph{Camera Viewpoint}, \emph{Language}, \emph{Sensor Noise}, and \emph{Robot Initial State}, as introduced in Section~\ref{subsec:generalization}. For each task-shift combination, we perform \newer{25} rollouts \reportchange{and adopt corresponding in-distribution demonstrations as training data.}

Table~\ref{tab:ood_results} summarizes the results \newer{across all eight HUMANO suites (100 tasks)}. \newer{GR00T N1.7 achieves the strongest overall OOD performance, with an average success rate of 20.0\%, followed by $\pi_{0.5}$ at 17.8\%. GR00T N1.7 leads in six of the seven OOD settings, while $\pi_{0.5}$ performs best under \emph{Camera Viewpoint} shifts (14.3\% vs. 11.0\%).} In contrast, Diffusion Policy, Flow Matching, and ACT achieve average success rates of only 2.5\%, \reportchange{5.7\%}, and 5.4\%, respectively, revealing a substantial robustness gap between the strongest generalist policies and conventional imitation-learning baselines.

The difficulty also varies across OOD settings. Averaged over all six baselines, \emph{Sensor Noise} is the most challenging shift, with a success rate of \newer{3.9\%, followed by \emph{Camera Viewpoint} at 5.9\%}, while \emph{Language} and \emph{Background} are less disruptive (\reportchange{14.7\% and 14.6\%}, respectively). The stronger performance under background variation may partly benefit from the scene diversity already present in the in-distribution training data. Overall, OOD robustness remains highly sensitive to the type of distribution shift, especially \newer{sensor noise and camera viewpoint changes}.

\subsection{Ablation and Analysis Study}
\label{subsec:ablation}

\begin{table*}[t]
\centering
\caption{
Effect of pretraining on HUMANO Basic suite.
We report in-distribution (ID) and out-of-distribution (OOD) success rates (\%). Ep. denotes training epoch.
\reportchange{Within each column, \textbf{bold} and \underline{underline} mark the two highest distinct positive scores, respectively.}
}
\label{tab:pretraining_ablation}
\vskip -0.05in
{\fontsize{8.5pt}{9pt}\selectfont
\setlength{\tabcolsep}{2.5pt}
\renewcommand{\arraystretch}{1.1}
\begin{tabular*}{\textwidth}{@{\extracolsep{\fill}}lccccccccccc}
\toprule
\multicolumn{3}{c}{Training Configuration} & \\
\cmidrule(lr){1-3}
Pretrain &
Pre. Ep. &
Basic Ep. &
ID &
Distr. &
Light. &
Backg. &
Camera &
Lang. &
Sensor &
Init. &
OOD Avg. \\
\midrule
None
& -- & 73
& \reportchange{\underline{21.6}} & \reportchange{\underline{17.2}} & \reportchange{\underline{16.0}} & \reportchange{\underline{24.8}}
& \reportchange{\underline{8.8}} & \reportchange{\underline{22.0}} & \reportchange{\underline{9.2}} & \reportchange{\underline{12.0}} & \reportchange{\underline{15.7}} \\

HUMANO
& 10 & 5
& \textbf{54.4} & \textbf{35.6}
& \textbf{48.8} & \textbf{62.4}
& \textbf{15.2} & \textbf{54.4}
& \textbf{29.6} & \textbf{43.6}
& \textbf{41.4} \\

SIMPLE
& 10 & 5
& 16.4 & 7.6 & 12.4 & 14.8
& 0.8 & 20.0 & 3.2 & 7.6 & 9.5 \\
\bottomrule
\end{tabular*}
}
\end{table*}

\begin{figure*}[t]
\centering
\includegraphics[width=0.9\textwidth]{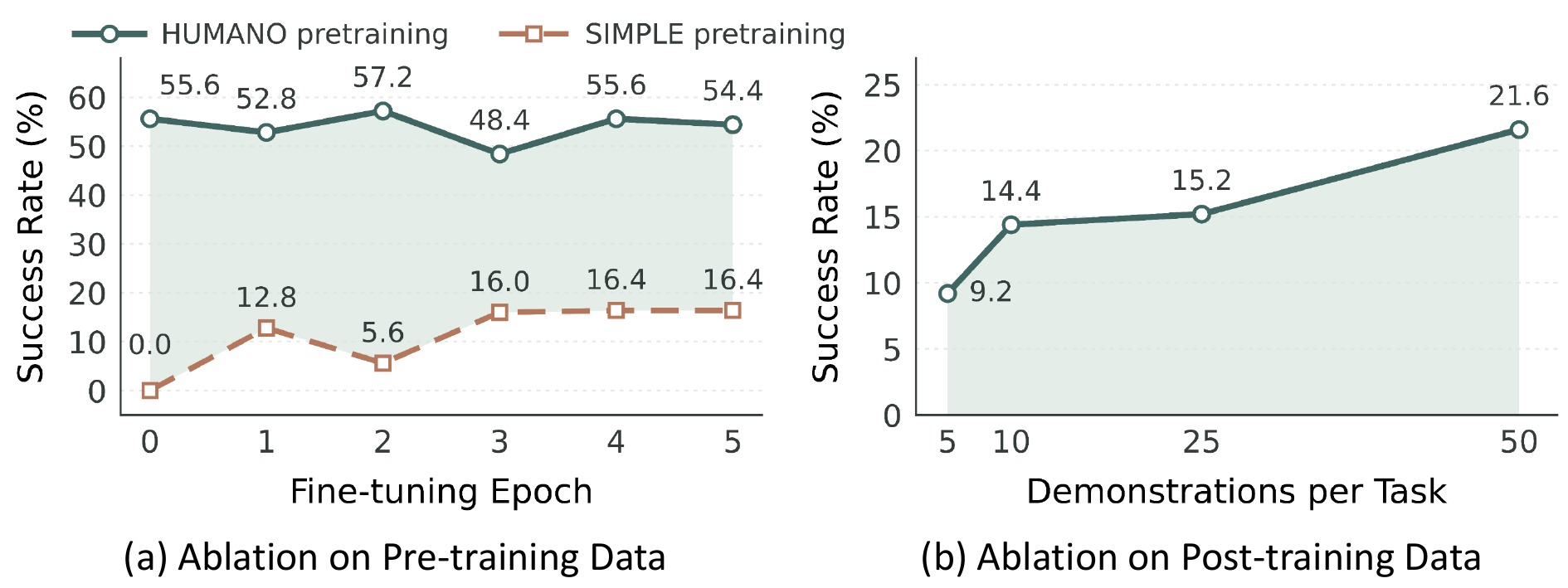}
\vskip -0.07in
\caption{
Effects of \newer{pre-training data} and post-training demonstration scale on HUMANO.
\newer{(a) Success rates across fine-tuning epochs following pre-training on HUMANO and SIMPLE, respectively.}
(b) Success rates with increasing post-training demonstration budgets per task.
}
\label{fig:scaling}
% \vskip -0.1in
\end{figure*}

\textbf{Ablation on Pre-training Data.}
We initialize GR00T-N1.7 and continue pretraining it for 10 epochs on either the eight HUMANO training suites or the SIMPLE dataset~\cite{wei2026simple}, followed by 5 epochs of post-training on the HUMANO Basic suite. Detailed training configurations are provided in Appendix~\ref{app:pretrain_details}.

Table~\ref{tab:pretraining_ablation} reports the impact of pre-training data on downstream performance. HUMANO pretraining achieves an ID success rate of 54.4\% and an OOD average of 41.4\%, substantially outperforming SIMPLE pretraining (16.4\% and 9.5\%) under the same training schedule. Figure~\ref{fig:scaling}(a) further shows that this advantage persists across all evaluated fine-tuning checkpoints, with HUMANO-pretrained policies consistently outperforming their SIMPLE-pretrained counterparts (48.4\%--57.2\% vs. 0.0\%--16.4\%). This suggests that the benefits of pre-training data are retained during downstream adaptation. Such improvements may stem from HUMANO's larger data scale and broader task diversity, which expose the model to richer visual observations and action distributions. Moreover, HUMANO pretraining followed by only five epochs of Basic-suite post-training achieves substantially higher performance than training from scratch for 73 epochs (54.4\% vs. 21.6\% ID and 41.4\% vs. 15.7\% OOD). Collectively, these results demonstrate the value of diverse multi-task pretraining for improving both downstream performance and robustness to distribution shifts.

\textbf{Ablation on Ratio of Post-training Data.}
We first study how multi-task performance scales with the amount of demonstration data. Specifically, we train the same policy using 10\%, 20\%, 50\%, and 100\% of the available demonstrations, while keeping the model architecture and training configuration fixed. 
As shown in Figure~\ref{fig:scaling}(b), increasing the demonstration budget from 5 to 50 per task improves the success rate from 9.2\% to 21.6\%, with non-uniform gains across data scales. 
These results demonstrate that increasing demonstration coverage consistently benefits multi-task policy learning, although the gains are not uniform across data scales. In particular, the improvement at the largest data budget suggests that additional demonstrations can still provide meaningful benefits.

\textbf{Single-Task vs. Multi-Task Learning.}
We compare single-task and multi-task learning under the same per-task demonstration budget. As shown in Table~\ref{tab:single_multi}, single-task training achieves 40.4\% average success, substantially outperforming multi-task training (21.6\%). It also performs better on eight of the ten tasks, while multi-task training leads only on Ball Kick; both fail on Table Button Press. These results suggest that joint training may introduce task interference rather than consistently enabling positive transfer, highlighting the challenge of balancing shared representations and task-specific capabilities in humanoid policy learning.

\begin{table*}[!t]
\centering
\fontsize{8pt}{9pt}\selectfont
\caption{Single-task versus suite-level training of GR00T N1.7 on the Basic suite. We report success rate (\%) for each task and the average across all 10 tasks.}
\label{tab:single_multi}
\vskip -0.05in
\setlength{\tabcolsep}{2pt}
\renewcommand{\arraystretch}{1.15}
\begin{tabular*}{\textwidth}{@{\extracolsep{\fill}}l*{11}{c}@{}}
\toprule
& \multicolumn{4}{c}{Motion Tasks}
& \multicolumn{3}{c}{Ground Picking}
& \multicolumn{3}{c}{Tabletop Tasks}
& \\
\cmidrule(lr){2-5}
\cmidrule(lr){6-8}
\cmidrule(lr){9-11}
Setting &
\makecell{Ball\\Kick} &
\makecell{Chair\\Push} &
\makecell{Chair\\Sit} &
\makecell{Dolly\\Push} &
\makecell{Basket\\Pick} &
\makecell{Bottle\\Pick} &
\makecell{Box\\Pick} &
\makecell{Apple\\Pick} &
\makecell{Button\\Press} &
\makecell{Plate\\Pick} &
Avg. \\
\midrule
Task-Level
& 0 & 56 & 16 & 96 & 56 & 20 & 40 & 56 & 0 & 64 & 40.4 \\

Suite-Level
& 40 & 12 & 0 & 68 & 48 & 0 & 20 & 16 & 0 & 12 & 21.6 \\
\bottomrule
\end{tabular*}
\end{table*}

\textbf{Generalization Study on Unseen Tasks.}
To further investigate generalization beyond the training task distribution, we conduct additional experiments on three held-out tasks that are not explicitly included in the HUMANO training set: \emph{Pear Picking}, \emph{Drawer Closing}, and \emph{Bottle Reorientation}. These tasks introduce different levels of novelty in object identity, manipulation goals, and required action sequences. We directly evaluate the GR00T N1.7 policy trained on the original HUMANO data, without any task-specific demonstrations or additional fine-tuning. Each task is evaluated over 25 rollouts.

Table~\ref{tab:heldout_tasks} reports the success rates on these held-out tasks. GR00T N1.7 achieves 8.0\% success on Pear Picking and 84.0\% on Drawer Closing, suggesting that previously learned manipulation skills can transfer to certain unseen tasks. In contrast, the policy achieves 0\% success on Bottle Reorientation, which requires grasping a horizontally placed bottle, reorienting it upright, and placing it stably on a table. These results indicate that current VLA policies exhibit a certain degree of generalization to unseen tasks, while generalizing to substantially novel manipulation skills remains challenging.

\begin{table}[t]
\centering
\small
\caption{
Generalization to three held-out tasks not explicitly included in the HUMANO training set. We report success rates over 25 evaluation rollouts.
}
\label{tab:heldout_tasks}
\vskip -0.05in
\begin{tabular}{llc}
\toprule
Task & Description & Success Rate (\%) \\
\midrule
Pear Picking & Pick up a pear & 8.0 \\
Drawer Closing & Close an open drawer & 84.0 \\
Bottle Reorientation & Reorient a lying bottle upright and place it stably & 0.0 \\
\bottomrule
\end{tabular}
\end{table}

\begin{table*}[t]
\centering
\small
\caption{
Simulation-to-real evaluation on four representative HUMANO tasks.
We report simulation success rate (Sim) and real-world success rate (Real), both in \%.
}
\label{tab:sim2real}
\vskip -0.05in
% \resizebox{\textwidth}{!}{
\begin{tabular}{lcccccccc}
\toprule
& \multicolumn{2}{c}{Apple Picking}
& \multicolumn{2}{c}{Chair Pushing}
& \multicolumn{2}{c}{Banana into Tray}
& \multicolumn{2}{c}{Basket Moving} \\
\cmidrule(lr){2-3}
\cmidrule(lr){4-5}
\cmidrule(lr){6-7}
\cmidrule(lr){8-9}
Method
& Sim & Real
& Sim & Real
& Sim & Real
& Sim & Real \\
\midrule
$\pi_{0.5}$
& 75 & 20 & 75 & 50 & 55 & 30 & 45 & 10 \\

$\Psi_0$
& 75 & 20 & 85 & 60 & 60 & 40 & 50 & 10 \\

GR00T N1.7
& 80 & 30
& 85 & 80
& 70 & 50
& 55 & 20 \\
\bottomrule
\end{tabular}
% }
\vskip -0.05in
\end{table*}

\begin{figure*}[!t]
\centering
\includegraphics[width=\textwidth]{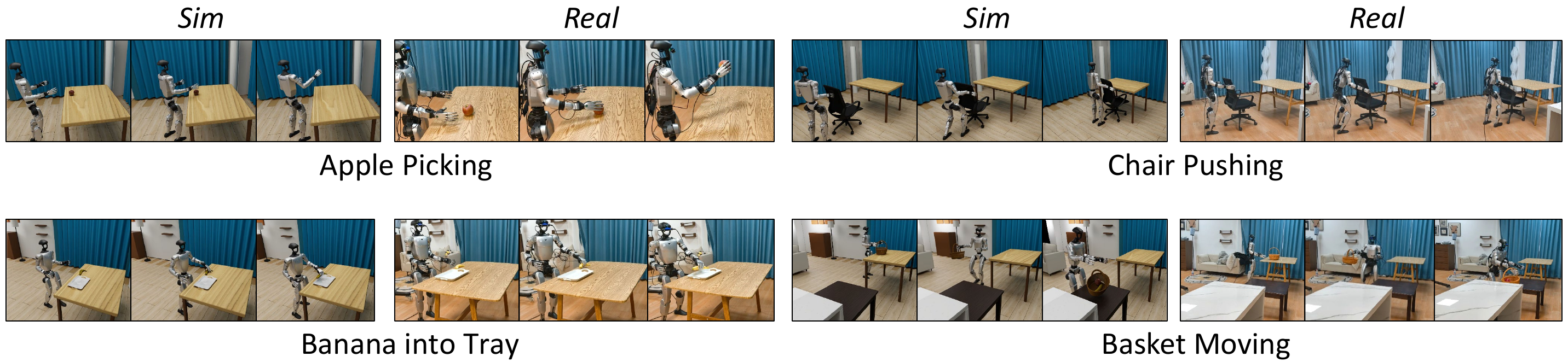}
\vskip -0.05in
\caption{\newer{Visualization of sim-to-real transfer results across four tasks.}}
\label{fig:sim2real}
\end{figure*}

\subsection{Sim-to-Real Evaluation}

\newer{
Finally, we evaluate sim-to-real transfer on four representative tasks: \emph{Apple Picking}, \emph{Chair Pushing}, \emph{Banana into Tray}, and \emph{Basket Moving}, using a Unitree G1 humanoid robot equipped with BrainCo Revo2 dexterous hands. For each task, we collect 40 simulation demonstrations and independently train three policies ($\pi_{0.5}$, $\Psi_0$, and GR00T N1.7) using the corresponding single-task data and default configurations. As shown in Figure~\ref{fig:sim2real} and Table~\ref{tab:sim2real}, each policy is evaluated with 20 simulation rollouts and 10 real-world trials per task. GR00T achieves the highest average success rates in both simulation and reality (72.5\% and 45.0\%, respectively), demonstrating the transferability of simulation-trained whole-body manipulation policies while revealing a remaining sim-to-real gap.
}
\section{Conclusion}
\label{sec:conclusion}

We introduced \textsc{HUMANO}, a large-scale benchmark for generalist humanoid loco-manipulation with 100 tasks across eight capability suites, 90 interactive scenes, and over 2K object assets. HUMANO provides controlled OOD evaluation and a unified SONIC-based motion-latent interface for standardized policy comparison. Experiments show promising multi-task capability in generalist robot policies, continued gains from increased demonstration scale, and preliminary sim-to-real transfer, while substantial capability and robustness gaps remain. We hope HUMANO serves as a scalable platform for advancing general and robust humanoid loco-manipulation.

\clearpage
\phantomsection
\section*{Authors}
\label{sec:authors}
{\raggedright\sffamily\setlength{\parskip}{0pt}\ReportAuthors\par}

\bibliographystyle{unsrtnat}
\bibliography{references}

% Keep a short appendix in the natural flow; insert \clearpage for a long report.
\clearpage

\appendix

In this appendix, we provide additional benchmark details, implementation clarifications, and experimental results:
\begin{itemize}
    \item Complete definitions of all 100 tasks in HUMANO across the eight task suites (see Section~\ref{app:tasks}).
    \item Detailed implementation protocols of the seven controlled OOD settings in HUMANO-Plus (see Section~\ref{app:ood_settings}).
    \item Training configurations of all evaluated baselines, together with the unified evaluation protocol (see Sections~\ref{app:baseline_implementation} and~\ref{app:baseline_evaluation}).
    \item Implementation details of the ablation studies (see Section~\ref{app:ablation_analysis}).
    \item Additional task-wise in-distribution results and suite-wise out-of-distribution results (see Section~\ref{app:additional_experiments}).
    \item A discussion of the current limitations of HUMANO (see Section~\ref{app:limitations}).
\end{itemize}

\section{Full Task Definitions}
\label{app:tasks}

This section provides the complete task definitions of HUMANO. The benchmark contains 100 tasks organized into eight suites: \emph{Basic}, \emph{Advanced}, \emph{Spatial}, \emph{Object}, \emph{Movement}, \emph{Goal}, \emph{Long-Horizon}, and \emph{Conditional}. These suites are designed to separate different sources of difficulty in humanoid loco-manipulation. Rather than constructing a large task set through minor scene or object perturbations, HUMANO varies task semantics, spatial configuration, manipulated objects, target goals, temporal composition, and environment-dependent decision rules in a controlled manner. For suites involving structured task variants, the tables below explicitly specify the corresponding object choices, target choices, or conditional branches.

\paragraph{Basic Suite.}
The Basic suite serves as the foundation for the remaining suites. It contains 10 tasks covering representative whole-body interaction primitives, including tabletop grasping, floor-level grasping, pushing, sitting, button pressing, and kicking. These tasks span different interaction heights and body configurations: tabletop tasks primarily require reaching and grasping, floor-level tasks involve larger posture changes, while pushing, sitting, and kicking require coordinated whole-body control and balance. Each task has a single primary objective, providing a basic assessment of fundamental humanoid interaction skills before additional generalization, composition, or conditional reasoning is introduced.

\paragraph{Advanced Suite.}
The Advanced suite extends the Basic suite from primitive execution to goal-directed loco-manipulation. Its 10 tasks require the robot not only to interact with an object, but also to move or transform it toward a specified target state, such as placing an object into a receptacle, moving a floor-level object onto a table, pushing furniture toward a destination, or pouring an object from a container. Compared with the Basic suite, these tasks require longer coordination between perception, locomotion, manipulation, and object placement while retaining relatively simple task semantics. The suite therefore evaluates whether individually acquired whole-body skills can be integrated into complete goal-directed behaviors.

\paragraph{Spatial Suite.}
The Spatial suite evaluates whether a policy can preserve the same task semantics under changes in spatial configuration. It contains 20 tasks derived from representative interaction patterns in the Basic and Advanced suites, while systematically varying object locations, interaction heights, robot-object relations, or target locations. For example, tabletop objects may appear at different corners, edges, or the center of a surface; grasp targets may be located on a high table, a low table, or the floor; and the robot may be initialized on different sides of the object. Other tasks use multiple spatial arrangements of the same source and target objects. By keeping the manipulation objective fixed while changing geometry, this suite separates spatial generalization from changes in object identity or task goal. \reportchange{Table~\ref{tab:tasks_spatial} lists the objective of each task.}

\begin{table*}[t]
\centering
\small
\caption{Full task definitions of the Basic and Advanced suites.}
\label{tab:tasks_basic_advanced}
\vskip 0.05in
\begin{tabular}{c p{0.37\textwidth} c p{0.35\textwidth}}
\toprule
\multicolumn{2}{c}{\textbf{Basic Suite}} &
\multicolumn{2}{c}{\textbf{Advanced Suite}} \\
\cmidrule(lr){1-2} \cmidrule(lr){3-4}
Task ID & Task Definition & Task ID & Task Definition \\
\midrule
B-1  & Pick up an apple from a table.
& A-1  & Put an apple into a basket. \\

B-2  & Pick up a plate from a table \reportchange{with both hands}.
& A-2  & Place a plate \reportchange{inside} a \reportchange{shelving unit}. \\

B-3  & Push a chair.
& A-3  & \reportchange{Place} a milk carton into a trash \reportchange{can}. \\

B-4  & Push a \reportchange{dolly}.
& A-4  & Put bread onto a tray. \\

B-5  & Sit on a chair.
& A-5  & Pick up a basket from the floor and place it on a table. \\

B-6  & Press a button.
& A-6  & \reportchange{Pick up} a box \reportchange{from the floor} and place it on a table. \\

B-7  & Pick up a bottle from the floor.
& A-7  & Pick up a bottle from the floor and place it on a table. \\

B-8  & Pick up a box from the floor.
& A-8  & Push a chair to the front of a table. \\

B-9  & Kick a ball.
& A-9  & Push a \reportchange{dolly until most of it is under} the table. \\

B-10 & Pick up a basket from the floor.
& A-10 & Pour an eraser out of a \reportchange{pen} holder \reportchange{onto a table}. \\
\bottomrule
\end{tabular}
\end{table*}

\begin{table*}[t]
\centering
\small
\caption{Full task definitions of the Spatial suite.}
\label{tab:tasks_spatial}
\vskip 0.05in
\begin{tabular}{c p{0.85\textwidth}}
\toprule
Task ID & Task Definition \\
\midrule
S-1  & Pick up an apple from \reportchange{a table}. \\
S-2  & Pick up a plate from \reportchange{a table with both hands}. \\
S-3  & Pick up a bottle from three heights: a high table, a low table, and the floor. \\
S-4  & Pick up a box from three heights: a high table, a low table, and the floor. \\
S-5  & \reportchange{Sit on a chair.} \\
S-6  & Push a \reportchange{dolly}. \\
S-7  & Push a chair. \\
S-8  & Kick a ball. \\
S-9  & Press a button. \\
S-10 & Pick up a basket from three heights: the floor, a low table, and a high table. \\
S-11 & Put an apple into a basket. \\
S-12 & Place a plate \reportchange{inside} a \reportchange{shelving unit}. \\
S-13 & \reportchange{Place} a milk carton into a trash \reportchange{can}. \\
S-14 & Put bread onto a tray. \\
S-15 & Pick up a bottle from the floor and place it on a table. \\
S-16 & Pick up a box from the floor and place it on a table. \\
S-17 & Push a chair to the front of a table. \\
S-18 & \reportchange{Push} a \reportchange{dolly until most of it is under} the table. \\
S-19 & Pour an eraser out of a \reportchange{pen} holder \reportchange{onto a table}. \\
S-20 & Pick up a basket from the floor and place it on a table. \\
\bottomrule
\end{tabular}
\end{table*}

\begin{table*}[t]
\centering
\small
\caption{Full task definitions of the Object and Movement suites.}
\label{tab:tasks_object_movement}
\vskip 0.05in
\begin{tabular}{c p{0.39\textwidth} c p{0.32\textwidth}}
\toprule
\multicolumn{2}{c}{\textbf{Object Suite}} &
\multicolumn{2}{c}{\textbf{Movement Suite}} \\
\cmidrule(lr){1-2} \cmidrule(lr){3-4}
Task ID & Task Definition & Task ID & Task Definition \\
\midrule
O-1  & Single-handed tabletop grasping of an apple, banana, or cake.
& M-1  & \reportchange{Walk to a table and pick} up an apple. \\

O-2  & Two-handed tabletop grasping of a plate or tray.
& M-2  & \reportchange{Walk to a table and pick} up a plate \reportchange{with both hands}. \\

O-3  & Pick up an apple \reportchange{or} banana \reportchange{with one hand}, or \reportchange{a plate or} tray \reportchange{with both hands,} from a table.
& M-3  & \reportchange{Walk to} a chair \reportchange{and push it}. \\

O-4  & Single-handed floor-level grasping of a bottle or basket.
& M-4  & \reportchange{Walk to} a \reportchange{dolly and push it}. \\

O-5  & Two-handed floor-level grasping of a box or storage box.
& M-5  & \reportchange{Walk to} a chair \reportchange{and sit down}. \\

O-6  & Pick up a bottle \reportchange{or basket with one hand}, \reportchange{or a} box or storage box \reportchange{with both hands,} from the floor.
& M-6  & \reportchange{Walk to a table and press} a button. \\

O-7  & Put a small object—an apple, banana, or cake—into a basket \reportchange{with one hand}.
& M-7  & \reportchange{Walk to} a bottle \reportchange{on} the floor \reportchange{and pick it up}. \\

O-8  & Put a large object—a plate or tray—\reportchange{upright inside} a cabinet \reportchange{with both hands}.
& M-8  & \reportchange{Walk to} a box \reportchange{on} the floor \reportchange{and pick it up}. \\

O-9  & Throw a tabletop object—milk \reportchange{carton}, bread, or cake—into a trash \reportchange{can}.
& M-9  & \reportchange{Walk to} a ball \reportchange{and kick it}. \\

O-10 & Put food—bread or cake—onto a tray.
& M-10 & \reportchange{Walk to} a basket \reportchange{on} the floor \reportchange{and pick it up}. \\
\bottomrule
\end{tabular}
\end{table*}

\paragraph{Object Suite.}
The Object suite focuses on generalization across manipulated objects while preserving the underlying interaction pattern. Each task defines a shared manipulation objective and instantiates it with multiple object categories or geometries. The suite includes both single-handed and two-handed grasping, tabletop and floor-level interaction, and goal-directed placement into baskets, cabinets, trays, or trash bins. For example, the same grasping behavior may be instantiated with an apple, banana, or cake, while larger-object tasks vary between plates, trays, boxes, and storage boxes. This construction tests whether a policy learns reusable manipulation behavior rather than overfitting to the visual appearance or geometry of a particular object instance.

\paragraph{Movement Suite.}
The Movement suite contains 10 tasks whose high-level objectives mirror the fundamental interaction patterns in the Basic suite, including grasping, pushing, sitting, pressing, and kicking. Keeping the task semantics aligned with the Basic suite provides a controlled basis for studying movement-related variation without simultaneously changing the underlying task objective.

\paragraph{Goal Suite.}
The Goal suite evaluates whether a policy can condition its behavior on the specified task goal when multiple valid interaction targets are simultaneously present. \reportchange{Across its 10 tasks, the instruction specifies either the destination for a given object or the object to place at a shared destination.} For example, an apple may need to be placed into a basket, box, or tray, while a plate may be assigned to a basket, box, or cabinet. Because the alternative targets remain available in the scene, successful execution requires the policy to identify the instructed goal rather than simply applying a fixed object-to-target association. This suite therefore isolates goal-conditioned generalization from changes in the basic manipulation skill itself.

\begin{table*}[t]
\centering
\small
\caption{Full task definitions of the Goal suite.}
\label{tab:tasks_goal}
\vskip 0.05in
% \resizebox{\textwidth}{!}{
\begin{tabular}{c p{0.85\textwidth}}
\toprule
Task ID & Task Definition \\
\midrule
G-1  & Put an apple \reportchange{at} the specified target\reportchange{: in} a basket, \reportchange{in a} box, \reportchange{or on a} tray. \\
G-2  & Put a plate into the specified target among a basket, box, and cabinet. \\
G-3  & Put bread \reportchange{at} the specified target\reportchange{: in} a basket, \reportchange{on a} tray, \reportchange{or in a} trash \reportchange{can}. \\
G-4  & Put \reportchange{a} milk \reportchange{carton} into the specified target among a basket, box, and trash \reportchange{can}. \\
G-5  & \reportchange{Put the specified object—an apple, banana, or bread—into a basket.} \\
G-6  & Put a book \reportchange{at} the specified target\reportchange{: upright on the lower shelf or inside} a cabinet. \\
G-7  & Pick up a bottle from the floor and place it \reportchange{at} the specified target\reportchange{: on} a table \reportchange{or inside a} cabinet. \\
G-8  & Pour an eraser from a \reportchange{pen} holder onto a table \reportchange{or into a} cabinet\reportchange{, as specified by the instruction}. \\
G-9  & \reportchange{Put the specified object—a book, floor-level bottle, or plate—into a storage box.} \\
G-10 & \reportchange{Move} a soccer ball \reportchange{from a table to} the specified target on the floor or \reportchange{inside} a cabinet. \\
\bottomrule
\end{tabular}
% }
\end{table*}

\paragraph{Long-Horizon Suite.}
The Long-Horizon suite contains 20 tasks that compose multiple locomotion and manipulation primitives into extended behavioral sequences. Unlike the Basic and Advanced suites, successful completion requires several intermediate states to be reached in the correct order. Representative tasks include opening a drawer before placing an object inside, placing an object onto a receptacle before lifting it, moving a container before inserting another object, and sequentially manipulating multiple objects. These tasks introduce longer temporal dependencies and expose policies to error accumulation: an early failure in grasping, navigation, placement, or articulated interaction can prevent all subsequent stages from being completed. The suite is therefore designed to evaluate temporal skill composition rather than isolated primitive success.

\begin{table*}[t]
\centering
\small
\caption{Full task definitions of the Long-Horizon suite.}
\label{tab:tasks_long}
\vskip 0.05in
% \resizebox{\textwidth}{!}{
\begin{tabular}{c p{0.85\textwidth}}
\toprule
Task ID & Task Definition \\
\midrule
L-1  & \reportchange{Open a drawer, pick} up an apple from a table, and place the apple inside. \\
L-2  & Open a drawer, pick up a plate from a table, and place the plate inside. \\
L-3  & Put both bread and cake onto a tray. \\
L-4  & Put bread onto a tray and then lift the tray. \\
L-5  & Put cake onto a plate and then lift the plate. \\
L-6  & Pick up a box from the floor, place it on a \reportchange{dolly}, and push the \reportchange{dolly} forward. \\
L-7  & Push a chair and then sit on it. \\
L-8  & Move a soccer ball from a table to the floor and then kick it. \\
L-9  & Move a basket from the floor to a table and then put a tabletop apple into the basket. \\
L-10 & \reportchange{Put a floor-level bottle into} a box \reportchange{on} a table and then \reportchange{lift} the box. \\
L-11 & Put both an apple and a banana into a basket. \\
L-12 & Move an apple from a basket onto a table and \reportchange{put} a banana into a trash \reportchange{can}. \\
L-13 & Pour two peanuts from a \reportchange{cup} onto a plate. \\
L-14 & \reportchange{Move a cake from a storage box onto a table and then put bread onto a tray.} \\
L-15 & Put an apple into a basket \reportchange{and then place} the basket on the floor. \\
L-16 & Move an apple from a high table into a drawer of a low table. \\
L-17 & Move a plate from a low table into a drawer of a high table. \\
L-18 & Move a \reportchange{box} away and then push a chair to the front of a table. \\
L-19 & \reportchange{Pick up} a book \reportchange{from} a \reportchange{high table and place it upright on a shelf}. \\
L-20 & Move a book from a \reportchange{shelf} into a tabletop drawer. \\
\bottomrule
\end{tabular}
% }
\end{table*}

\paragraph{Conditional Suite.}
The Conditional suite evaluates state-dependent decision making. Each of its 10 tasks admits more than one valid action sequence, and the correct sequence is determined by the observed environment state. Some tasks explicitly require branching on articulated-object state, such as opening a drawer only when it is closed. Others require selecting an empty container, choosing a basket according to its contents, or deciding whether an obstructing object must first be moved. Consequently, success cannot be achieved by executing a single fixed trajectory for every episode; the policy must first identify the relevant state and then select the corresponding action sequence. This suite complements the Long-Horizon suite by introducing conditional branching in addition to temporal composition.

\begin{table*}[t]
\centering
\small
\caption{Full task definitions of the Conditional suite.}
\label{tab:tasks_conditional}
\vskip 0.05in
\begin{tabular}{c p{0.88\textwidth}}
\toprule
Task ID & Task Definition \\
\midrule
C-1  & If the \reportchange{drawer} is closed, open it and put the plate inside; otherwise, put the plate inside. \\
C-2  & If the drawer is closed, open it and put the \reportchange{apple} inside; otherwise, put the \reportchange{apple} inside. \\
C-3  & \reportchange{If the left storage box is empty, put the} bottle \reportchange{inside; otherwise, put it} into the \reportchange{right storage box}. \\
C-4  & \reportchange{If} the \reportchange{left} basket contains no other fruit\reportchange{, put the apple inside; otherwise, put it into the right basket}. \\
C-5  & If the chair is not in front of the table, push it to the table and sit down; otherwise, sit down. \\
C-6  & If the basket is on the floor, move it onto the table and then put the apple inside; otherwise, put the apple inside. \\
C-7  & \reportchange{If the cake is on the tray, pour both} peanuts \reportchange{from} the \reportchange{cup onto} the plate\reportchange{; if the cake is on the plate, pour them onto the} tray. \\
C-8  & \reportchange{If the cake is on the tray, put the} bread \reportchange{onto} the plate\reportchange{; if the cake is on the plate, put the bread onto the} tray. \\
C-9  & If \reportchange{the dolly blocks the path to} the table, move it \reportchange{aside and push} the \reportchange{chair toward the table}; otherwise, push the \reportchange{chair toward the table}. \\
C-10 & If the \reportchange{second shelf} contains a book, place the tabletop book on the \reportchange{second shelf}; if the \reportchange{third shelf} contains a book, place it on the \reportchange{third shelf}. \\
\bottomrule
\end{tabular}
\end{table*}

\section{OOD Setting Details}
\label{app:ood_settings}

\subsection{Protocol Overview}

This section provides the implementation details of the seven controlled OOD settings introduced in Section~\ref{subsec:generalization}: \emph{Distractor}, \emph{Lighting}, \emph{Background}, \emph{Camera Viewpoint}, \emph{Language}, \emph{Sensor Noise}, and \emph{Robot Initial State}. The settings are designed as factorized diagnostic tests and are evaluated independently rather than composed within the same episode.

Except for \emph{Background}, each OOD episode is constructed from a corresponding in-distribution configuration by modifying only the designated factor. The task objective, success criterion, task-relevant objects, and all other environment states are kept unchanged. Sampled perturbation parameters and simulator states are stored and reused across evaluated policies, ensuring matched evaluation conditions. The \emph{Background} setting instead evaluates transfer to globally held-out scenes and therefore changes the underlying scene layout and appearance while preserving the task semantics.

\subsection{OOD Settings}

\paragraph{Distractor.}

The Distractor condition evaluates robustness to task-irrelevant visual and
physical clutter by introducing \reportchange{five} additional objects near the
interaction region
\citep{wei2026simple,chen2026robotwin,fei2026liberoplus}. To preserve the
original task semantics, objects that are semantically identical or closely
related to the manipulation targets, destination containers, and other
task-relevant entities are excluded from the candidate pool. Eligible semantic
categories are sampled before individual assets to reduce sampling bias toward
categories containing more object instances.

Distractors are initialized from stable poses and placed on support surfaces
already present in the task. Candidate placements are checked for physical
support, collision avoidance, scene-boundary compliance, and \reportchange{placement within
the head-mounted camera view}. The task-relevant objects, robot initial state, task
objective, and success criterion remain unchanged. Accepted object identities,
poses, and sampling seeds are stored for deterministic replay.

\paragraph{Lighting.}

The \emph{Lighting} condition evaluates robustness to illumination changes
while preserving the scene geometry, object configuration, robot state, and
task objective. Related benchmarks consider variations in light color,
intensity, position, direction, and shadow configuration
\citep{wei2026simple,chen2026robotwin,fei2026liberoplus}.

The ID configuration contains six cylindrical lights arranged in two rows of three, with color temperature sampled from $[2001,8001]\,\mathrm{K}$ and light radius sampled from $[0.08,0.12]\,\mathrm{m}$. OOD episodes instantiate one of three lighting subtypes: \reportchange{\emph{RGB color} alters the color of light sources; \emph{Local illumination} reduces the number of active lights; and \emph{Intensity} decreases the overall lighting intensity. Scene-authored lights are disabled, while light radii remain within their ID range.}%\emph{RGB color} applies a \reportchange{green-, magenta-, cyan-, blue-, red-, or amber-dominant} emission profile to all lights; \emph{Local illumination} \reportchange{keeps only one or two lights active} without compensating for the reduced illumination; and \reportchange{\emph{Intensity} reduces the intensity of all six lights from the nominal value of 10,000 to a value sampled from $[1000,2500]$}. 
\reportchange{Scene-authored lights are disabled, and light radii retain their ID range.} Only one subtype is applied per episode, and the resulting lighting configuration remains fixed throughout the rollout.

\paragraph{Background.}

Existing benchmarks define background variation at different granularities.
RoboTwin~2.0 and LIBERO-Plus primarily perturb scene textures and surface
materials within fixed task environments, whereas SIMPLE exposes policies to
diverse complete indoor scenes but does not impose a globally disjoint
room-level split for evaluating the same tasks
\citep{wei2026simple,chen2026robotwin,fei2026liberoplus}. HUMANO instead
defines \emph{Background} OOD using FloorPlans excluded from all training data,
thereby evaluating task transfer to unseen room appearance, geometry, furniture
configurations, and obstacle layouts.

The global scene split covers kitchens, living rooms, and bedrooms. For each
room category, 20 FloorPlans are assigned to the training pool and 10 to the
held-out pool, yielding 60 training and 30 held-out FloorPlans. All scene
variants derived from the same underlying FloorPlan remain on the same side of
the split, preventing an evaluation layout from appearing in the training data
of another task. Unlike the other OOD conditions, which are derived from
corresponding ID configurations, Background configurations are matched at the
task level and instantiated exclusively in held-out FloorPlans. The task
definition, task-relevant objects, objective, and success criterion are
preserved.

To ground a task in a candidate FloorPlan, HUMANO constructs a protected
spatial footprint from the training demonstration trajectories. For
table-based tasks, a compatible table or counter in the background scene is
used as a spatial anchor when available; the background anchor is subsequently
hidden so that the task-native support and object configuration remain
unchanged. Otherwise, the task is placed in a valid free-floor region. The
background is aligned using only planar translation and yaw, without scaling or
modifying the task geometry. A candidate placement is rejected if the complete
task footprint falls outside the room or intersects immutable scene structures.
Removable background assets that conflict with the task workspace or trajectory
are hidden together with their linked components and dependent supported
objects, avoiding partial furniture and unsupported scene fragments. The
selected alignment and asset-visibility state are recorded for deterministic
replay.

For tasks containing a task table, Background additionally samples from
material families disjoint from those used for training and ID evaluation.
These appearance changes do not modify the table's collision geometry or
physical properties. Because the held-out FloorPlan and table appearance may
change jointly, Background is evaluated as a compound scene-level shift rather
than a texture-only perturbation; its results should therefore not be
interpreted as isolating the effect of room geometry or appearance alone.

\paragraph{Camera Viewpoint.}

Camera-pose variation has been considered in both manipulation robustness
benchmarks and humanoid sim-to-real data generation. LIBERO-Plus perturbs the
position, orientation, and field of view of an external scene camera, whereas
OASIS randomizes camera extrinsics when rendering G1 training data
\citep{fei2026liberoplus,yu2026oasis}. HUMANO instead evaluates perturbations
of the egocentric head-mounted stereo rig while keeping the robot state,
physical task layout, camera intrinsics, and image resolution unchanged.

\reportchange{Humanoid-camera settings in OASIS and NICO motivate our evaluation of
viewpoint variation} \citep{yu2026oasis,gade2022simtoreal}.
OASIS applies camera randomization as a training
augmentation, and NICO changes the robot's head joints, whereas HUMANO applies
evaluation-time offsets to the camera-to-head extrinsic transformation.

\reportchange{Each Camera Viewpoint episode jointly perturbs yaw and pitch, with
absolute offsets in the range $[10^\circ,20^\circ]$:}
\begin{align}
\reportchange{\Delta\mathrm{yaw}} &\reportchange{{}= s_y a_y, \qquad a_y\in[10^\circ,20^\circ],}\\
\reportchange{\Delta\mathrm{pitch}} &\reportchange{{}= s_p a_p, \qquad a_p\in[10^\circ,20^\circ],}
\end{align}

\reportchange{where $s_y,s_p\in\{-1,+1\}$. Offsets are stratified by magnitude and
sign, covering the four combinations of yaw and pitch directions.
Translation and roll remain unchanged.} The
rotation is composed using static XYZ roll--pitch--yaw angles.
\reportchange{The joint perturbation changes both the horizontal and vertical
viewing directions while preserving the physical task configuration.}

The sampled transformation is applied to the stereo rig before deriving the
left- and right-eye poses. Consequently, both cameras undergo the same rigid
extrinsic perturbation while their relative orientation and stereo baseline
remain unchanged. The perturbed configuration is fixed throughout the episode
and stored with the episode state for deterministic replay.

\paragraph{Language.}
Following prior robustness benchmarks that vary linguistic expressions while
preserving task intent \citep{chen2026robotwin,fei2026liberoplus}, the
\emph{Language} setting \reportchange{constructs evaluation instructions from the
source instructions associated with each task's goals and conditions}.
\reportchange{These reformulations are held out from the HUMANO training annotations
and follow three styles}:

\begin{itemize}
    \item \textbf{T1---Action-oriented:} a direct description of the required
    action and intended outcome;
    \item \textbf{T2---Context-oriented:} a description that first establishes
    a task-relevant scene or object relation and then states the required
    action; and
    \item \textbf{T3---Paraphrased:} a semantically equivalent reformulation of
    the \reportchange{source} instruction using alternative wording or sentence structure.
\end{itemize}

\reportchange{The reformulations vary in wording and the amount of descriptive detail.} All variants preserve the task-relevant entities and
relations, subgoal order, terminal condition, and hand specification when
explicitly required by the task. They do not disclose sampled initial states,
numerical success thresholds, unobserved scene properties, or additional
execution strategies.

Only the instruction is replaced relative to the corresponding ID
configuration. The \reportchange{instruction} variants are constructed and stored offline and are
assigned deterministically across evaluation episodes \reportchange{according to the
stored episode configurations}.

\paragraph{Sensor Noise.}
Following the Gaussian-blur evaluation in LIBERO-Plus
\citep{fei2026liberoplus}, the \emph{Sensor Noise} setting applies Gaussian
blur to both RGB views of the head-mounted stereo camera. The blur standard
deviation is \reportchange{selected} from the discrete set

\begin{equation}
    \sigma \in \{1,2,\ldots,10\}\ \mathrm{pixels}.
\end{equation}

The same severity is used for both views and remains fixed throughout an
episode. Blur is applied independently to the RGB channels before any
policy-specific resizing or normalization. All non-RGB observations and the
underlying simulator state remain unchanged. Whereas LIBERO-Plus applies the
corruption to its agent-view observation, HUMANO applies it consistently to
both egocentric head-camera views. Thus, this condition specifically measures
robustness to controlled RGB blur rather than to general sensor corruption.

\paragraph{Robot Initial State.}
Prior benchmarks evaluate initial-state robustness through variations in
manipulator starting poses or task-specific humanoid initialization regions
\citep{fei2026liberoplus,wei2026simple}. HUMANO isolates changes in the
humanoid floating-base configuration while preserving the task objects and
non-root joint states. We consider two subtypes:

\begin{enumerate}
    \item \textbf{Translation.}
    A planar offset is sampled \reportchange{with a distance uniformly drawn from
    $[0.10,0.15]\,\mathrm{m}$ and a uniformly sampled direction}:
    \begin{align}
        \reportchange{r} &\reportchange{{}\sim\mathcal{U}(0.10,0.15)\ \mathrm{m},}\\
        \theta &\sim\mathcal{U}(0,2\pi),\\
        (\Delta x,\Delta y) &= (r\cos\theta,r\sin\theta).
    \end{align}

    \item \textbf{Yaw.}
    The base position is retained, and its heading is perturbed \reportchange{in either
    direction using the nominal range}
    \begin{equation}
        \reportchange{\Delta\psi=s a, \qquad a\sim\mathcal{U}(10^\circ,15^\circ),\quad s\in\{-1,+1\}.}
    \end{equation}
\end{enumerate}

Translation and yaw are evaluated separately. \reportchange{The yaw range is adjusted
to the spatial constraints and initial configuration of each task.} The base height, non-root joint
states, task-object poses, and remaining scene state are inherited from the
corresponding ID configuration. A sampled pose is accepted only if the robot
remains within the valid scene region, satisfies its joint limits, introduces
no new or deeper self-collision\reportchange{, avoids non-ground external contact}, and retains
\reportchange{foot--ground contact or the baseline foot--ground gap} without non-foot ground contact. Up to 100 candidates are
sampled; configurations for which no valid candidate is found are rejected
rather than replaced by the original ID pose.

For each accepted candidate, the complete simulator state is stored and reused
for deterministic evaluation. The validity test \reportchange{checks the initial
configuration, with additional startup-transition checks for selected tasks,}
and does not use policy success.

\section{Implementation Details of Baselines}
\label{app:baseline_implementation}

\subsection{Common Training Setup}

We evaluate six representative policies on the eight HUMANO suites: ACT~\citep{zhao2023learning}, Diffusion Policy~\citep{chi2025diffusion}, Flow Matching~\citep{bostondynamics2025largebehaviormodelsatlas}, $\pi_{0.5}$~\citep{intelligence2025pi05}, GR00T N1.7~\citep{bjorck2025gr00t}, and $\Psi_0$~\citep{wei2026psi0}. For each baseline, we train a separate policy for each suite using all training demonstrations from the corresponding tasks. All baselines share the same suite splits, observation and action interfaces, and optimization-step budget, while retaining their native architectures, preprocessing pipelines, and training objectives.

All experiments use eight Nvidia A100 GPUs with a per-GPU batch size of 64, resulting in a global batch size of 512 without gradient accumulation. Each model is trained for 20,000 optimization steps. ACT, Diffusion Policy, Flow Matching, and $\pi_{0.5}$ use the LeRobot v0.6.0 codebase, whereas GR00T N1.7 and $\Psi_0$ use their respective native implementations with HUMANO-specific data and embodiment adapters.

Each observation contains an egocentric RGB image,
$I_t \in [0,255]^{360 \times 640 \times 3}$,
and a 46-dimensional proprioceptive state,
\begin{equation}
s_t =
\left[
s_t^{\mathrm{joint}},
g_t^{\mathrm{proj}}
\right]
\in \mathbb{R}^{46},
\end{equation}
where $s_t^{\mathrm{joint}}\in\mathbb{R}^{43}$ denotes the joint state and $g_t^{\mathrm{proj}}\in\mathbb{R}^{3}$ is the projected gravity vector. Images are stored at their original resolution and processed using each model's native visual preprocessing pipeline.

All policies predict the same 78-dimensional action representation,
\begin{equation}
a_t =
\left[
a_t^{\mathrm{motion}},
a_t^{\mathrm{left}},
a_t^{\mathrm{right}}
\right]
\in \mathbb{R}^{78},
\end{equation}
where $a_t^{\mathrm{motion}}\in\mathbb{R}^{64}$ is the SONIC motion token and
$a_t^{\mathrm{left}},a_t^{\mathrm{right}}\in\mathbb{R}^{7}$ are the left- and right-hand commands. All baselines use one observation timestep and predict an action chunk of length $H=40$. Language-conditioned policies additionally receive the corresponding task instruction.

Training data are drawn exclusively from the HUMANO training split. For LeRobot-based baselines, the source datasets are converted to the LeRobot v3.0 format with explicit \texttt{observation.state} and \texttt{action} fields. GR00T and $\Psi_0$ use model-specific adapters that expose the same 46-dimensional state and 78-dimensional action interface to their native data loaders. Normalization statistics are computed from the training data, while model-specific normalization, padding, and validity masks are handled by the corresponding native processors.

Table~\ref{tab:baseline_implementation} summarizes the main training configurations of the six baselines.

\begin{table}[htbp]
\centering
\small
\caption{
Baseline training configurations. GR00T uses 40 valid action positions within its internally padded action representation. The $\pi_{0.5}$ learning rate follows the configured pretrained checkpoint.
}
\label{tab:baseline_implementation}
\vskip 0.05in
\begin{tabular}{lccc}
\toprule
Policy & Obs. Horizon & Action Horizon & Learning Rate \\
\midrule
ACT              & 1 & 40 & $1 \times 10^{-5}$ \\
Diffusion Policy & 1 & 40 & $1 \times 10^{-4}$ \\
Flow Matching    & 1 & 40 & $1 \times 10^{-4}$ \\
$\pi_{0.5}$      & 1 & 40 & $2 \times 10^{-4}$ \\
GR00T N1.7       & 1 & 40 & $2 \times 10^{-4}$ \\
$\Psi_0$         & 1 & 40 & $2 \times 10^{-4}$ \\
\bottomrule
\end{tabular}
\end{table}

\subsection{Baseline Configurations}

\paragraph{ACT.}
We use the LeRobot \texttt{ACTPolicy} implementation with an ImageNet-pretrained ResNet-18 visual encoder and a Transformer encoder--decoder. The Transformer uses a hidden dimension of 512, eight attention heads, four encoder layers, one decoder layer, a feed-forward dimension of 3,200, and dropout of 0.1. We retain the conditional variational formulation with a 32-dimensional latent variable and a four-layer latent encoder. Training minimizes an $\ell_1$ action reconstruction loss together with a KL regularization term weighted by 10. We set \texttt{n\_obs\_steps=1}, \texttt{chunk\_size=40}, and \texttt{n\_action\_steps=40}, with the state projection and action head adapted to the 46-dimensional state and 78-dimensional action spaces.

\paragraph{Diffusion Policy.}
Diffusion Policy uses an ImageNet-pretrained ResNet-18 visual encoder and a conditional one-dimensional U-Net. Visual features and proprioceptive states form the global condition for action-sequence denoising. The U-Net uses channel widths of 512, 1,024, and 2,048, a kernel size of five, and a 128-dimensional diffusion-timestep embedding. Training follows the DDPM noise-prediction objective with 100 diffusion timesteps and a squared-cosine noise schedule. We use one observation timestep, an action horizon of 40, and 10 reverse-diffusion steps at inference. States and actions are normalized using min--max statistics.

\paragraph{Flow Matching.}
The Flow Matching baseline uses LeRobot's \texttt{MultiTaskDiTPolicy} with the \texttt{flow\_matching} objective. Pretrained CLIP ViT-B/16 encoders provide visual and language features, which condition the action Transformer together with the proprioceptive state. The Transformer contains six layers with a hidden dimension of 512, eight attention heads, rotary positional embeddings, and dropout of 0.1; the CLIP text encoder remains frozen. The model learns the velocity field between Gaussian noise and the target action chunk using the native beta timestep sampler with $\alpha=1.5$, $\beta=1.0$, and $s=0.999$. Images are processed at $224\times224$, and inference uses 100-step Euler integration to generate a 40-step action sequence.

\paragraph{$\pi_{0.5}$.}
We use the LeRobot \texttt{PI05Policy} implementation initialized from a pretrained $\pi_{0.5}$ checkpoint. The PaliGemma vision-language backbone and Gemma action expert retain their native flow-matching objective, tokenizer, and prompt construction. The vision-language backbone is frozen, while the action expert and action projection layers are fine-tuned. We configure \texttt{max\_state\_dim=46}, \texttt{max\_action\_dim=78}, \texttt{chunk\_size=40}, and \texttt{n\_action\_steps=40}. Proprioceptive states are discretized using the native state tokenizer and incorporated into the language prompt, while images are resized with padding to $224\times224$. Quantile-based state and action normalization is retained.

The pretrained action interface has 32 dimensions, whereas HUMANO uses 78-dimensional actions. We therefore expand the action input projection, output projection, and output bias to the required dimensionality. Existing pretrained entries are preserved, while additional dimensions are initialized by cyclically copying the pretrained entries along the expanded dimension.

\paragraph{GR00T N1.7.}
GR00T N1.7 uses a Cosmos-Reason2-2B vision-language backbone followed by an embodiment-conditioned flow-matching action head. We initialize from the pretrained N1.7-3B checkpoint and register the \texttt{UNITREE\_G1\_SONIC} embodiment. The 46-dimensional state is organized into leg, waist, arm, hand, and projected-gravity fields, while the 78-dimensional action is divided into the SONIC motion token and bilateral hand commands. The native processor places these fields within GR00T's internally padded 132-dimensional state and action representations and constructs the corresponding validity masks. The action modality contains 40 valid future timesteps.

Fine-tuning updates the projectors and action head while keeping the language and visual backbones frozen. We use AdamW with a learning rate of $2\times10^{-4}$, a 5\% warm-up ratio, and weight decay of $10^{-5}$. Color jitter uses brightness 0.3, contrast 0.4, saturation 0.5, and hue 0.08.

\paragraph{$\Psi_0$.}
$\Psi_0$ uses a Qwen3-VL-2B-based vision-language backbone and an ActionTransformer action head. The action head contains six Transformer blocks with a hidden dimension of 1,536, 24 attention heads, and a feed-forward dimension of 2,048. We configure a 46-dimensional state, a 78-dimensional action, one observation timestep, and a 40-step action chunk. Training uses the native flow objective with 1,000 training timesteps, while inference uses 10 sampling steps. The vision-language backbone remains frozen and only the action head is optimized.

Pretrained action-head tensors with incompatible dimensions are loaded by copying the overlapping entries, while newly introduced entries retain their default initialization. Training uses bfloat16 mixed precision, gradient checkpointing, and AdamW with a learning rate of $2\times10^{-4}$, 1,000 warm-up steps, cosine decay to $5\times10^{-7}$, and weight decay of $10^{-8}$. We use the native state and action normalization schemes and image augmentation at $240\times320$ resolution.

% ==================== Basic ====================

\begin{table}[t]
\centering
\small
\caption{Per-task in-distribution success rates (\%) on the HUMANO Basic suite, evaluated over 25 rollouts per task. \reportchange{Within each row, \textbf{bold} and \underline{underline} mark the two highest distinct positive scores, respectively.}}
\label{tab:basic_tasks}
\setlength{\tabcolsep}{5pt}
\renewcommand{\arraystretch}{1.1}
\vskip -0.05in
\begin{tabular}{@{}>{\raggedright\arraybackslash}p{\dimexpr0.42\linewidth+12pt\relax}>{\centering\arraybackslash}p{40pt}>{\centering\arraybackslash}p{40pt}>{\centering\arraybackslash}p{23pt}>{\centering\arraybackslash}p{23pt}>{\centering\arraybackslash}p{30pt}>{\centering\arraybackslash}p{23pt}@{}}
\toprule
Task & \shortstack{Diffusion\\Policy} & \shortstack{Flow\\Matching} & ACT & $\pi_{0.5}$ & \shortstack{GR00T\\N1.7} & $\Psi_0$ \\
\midrule
\hyperref[tab:tasks_basic_advanced]{\reportchange{B-1}}\reportchange{:}\ Table\allowbreak{}Apple\allowbreak{}Pick & 0.0 & 0.0 & \underline{4.0} & \textbf{16.0} & \textbf{16.0} & 0.0 \\
\hyperref[tab:tasks_basic_advanced]{\reportchange{B-2}}\reportchange{:}\ Table\allowbreak{}Plate\allowbreak{}Pick & 0.0 & 0.0 & 0.0 & \textbf{52.0} & \underline{12.0} & 0.0 \\
\hyperref[tab:tasks_basic_advanced]{\reportchange{B-3}}\reportchange{:}\ Chair\allowbreak{}Push & 0.0 & \underline{4.0} & 0.0 & 0.0 & \textbf{12.0} & 0.0 \\
\hyperref[tab:tasks_basic_advanced]{\reportchange{B-4}}\reportchange{:}\ Dolly\allowbreak{}Push & 28.0 & 0.0 & 0.0 & \textbf{84.0} & \underline{68.0} & 36.0 \\
\hyperref[tab:tasks_basic_advanced]{\reportchange{B-5}}\reportchange{:}\ Chair\allowbreak{}Sit & 0.0 & 0.0 & 0.0 & 0.0 & 0.0 & 0.0 \\
\hyperref[tab:tasks_basic_advanced]{\reportchange{B-6}}\reportchange{:}\ Table\allowbreak{}Button\allowbreak{}Press & 0.0 & 0.0 & 0.0 & 0.0 & 0.0 & \textbf{4.0} \\
\hyperref[tab:tasks_basic_advanced]{\reportchange{B-7}}\reportchange{:}\ Ground\allowbreak{}Bottle\allowbreak{}Pick & 0.0 & 0.0 & 0.0 & 0.0 & 0.0 & \textbf{4.0} \\
\hyperref[tab:tasks_basic_advanced]{\reportchange{B-8}}\reportchange{:}\ Ground\allowbreak{}Box\allowbreak{}Pick & 0.0 & 0.0 & \underline{36.0} & \textbf{40.0} & 20.0 & 8.0 \\
\hyperref[tab:tasks_basic_advanced]{\reportchange{B-9}}\reportchange{:}\ Ball\allowbreak{}Kick & 0.0 & 0.0 & 0.0 & 8.0 & \textbf{40.0} & \underline{16.0} \\
\hyperref[tab:tasks_basic_advanced]{\reportchange{B-10}}\reportchange{:}\ Ground\allowbreak{}Basket\allowbreak{}Pick & 0.0 & 16.0 & 0.0 & 4.0 & \textbf{48.0} & \underline{20.0} \\
\midrule
Avg. & 2.8 & 2.0 & 4.0 & \underline{20.4} & \textbf{21.6} & 8.8 \\
\bottomrule
\end{tabular}
\end{table}

% ==================== Advanced ====================

\begin{table}[t]
\centering
\small
\caption{Per-task in-distribution success rates (\%) on the HUMANO Advanced suite, evaluated over 25 rollouts per task. \reportchange{Within each row, \textbf{bold} and \underline{underline} mark the two highest distinct positive scores, respectively.}}
\label{tab:advanced_tasks}
\setlength{\tabcolsep}{5pt}
\renewcommand{\arraystretch}{1.1}
\vskip -0.05in
\begin{tabular}{@{}>{\raggedright\arraybackslash}p{\dimexpr0.42\linewidth+12pt\relax}>{\centering\arraybackslash}p{40pt}>{\centering\arraybackslash}p{40pt}>{\centering\arraybackslash}p{23pt}>{\centering\arraybackslash}p{23pt}>{\centering\arraybackslash}p{30pt}>{\centering\arraybackslash}p{23pt}@{}}
\toprule
Task & \shortstack{Diffusion\\Policy} & \shortstack{Flow\\Matching} & ACT & $\pi_{0.5}$ & \shortstack{GR00T\\N1.7} & $\Psi_0$ \\
\midrule
\hyperref[tab:tasks_basic_advanced]{\reportchange{A-1}}\reportchange{:}\ Table\allowbreak{}Apple\allowbreak{}To\allowbreak{}Basket & 0.0 & 4.0 & 0.0 & \textbf{40.0} & \underline{28.0} & 8.0 \\
\hyperref[tab:tasks_basic_advanced]{\reportchange{A-2}}\reportchange{:}\ Table\allowbreak{}Plate\allowbreak{}To\allowbreak{}Shelf & 0.0 & 0.0 & 0.0 & \underline{12.0} & \textbf{48.0} & \underline{12.0} \\
\hyperref[tab:tasks_basic_advanced]{\reportchange{A-3}}\reportchange{:}\ Table\allowbreak{}Milk\allowbreak{}To\allowbreak{}Trash\allowbreak{}Can & 0.0 & 0.0 & 0.0 & \underline{4.0} & \textbf{8.0} & \underline{4.0} \\
\hyperref[tab:tasks_basic_advanced]{\reportchange{A-4}}\reportchange{:}\ Table\allowbreak{}Bread\allowbreak{}To\allowbreak{}Tray & 0.0 & 0.0 & 0.0 & 0.0 & \textbf{24.0} & 0.0 \\
\hyperref[tab:tasks_basic_advanced]{\reportchange{A-5}}\reportchange{:}\ Ground\allowbreak{}Basket\allowbreak{}To\allowbreak{}Table & 4.0 & 8.0 & 0.0 & \textbf{72.0} & 28.0 & \underline{32.0} \\
\hyperref[tab:tasks_basic_advanced]{\reportchange{A-6}}\reportchange{:}\ Ground\allowbreak{}Box\allowbreak{}To\allowbreak{}Table & 0.0 & 0.0 & 16.0 & \textbf{64.0} & \underline{52.0} & 4.0 \\
\hyperref[tab:tasks_basic_advanced]{\reportchange{A-7}}\reportchange{:}\ Ground\allowbreak{}Bottle\allowbreak{}To\allowbreak{}Table & 0.0 & 0.0 & 0.0 & \textbf{12.0} & \underline{4.0} & \underline{4.0} \\
\hyperref[tab:tasks_basic_advanced]{\reportchange{A-8}}\reportchange{:}\ Chair\allowbreak{}Push\allowbreak{}To\allowbreak{}Table & 0.0 & 0.0 & 4.0 & \underline{64.0} & \textbf{68.0} & \underline{64.0} \\
\hyperref[tab:tasks_basic_advanced]{\reportchange{A-9}}\reportchange{:}\ Dolly\allowbreak{}Push\allowbreak{}To\allowbreak{}Table & 0.0 & 0.0 & 4.0 & \textbf{88.0} & \underline{80.0} & 48.0 \\
\hyperref[tab:tasks_basic_advanced]{\reportchange{A-10}}\reportchange{:}\ Table\allowbreak{}Eraser\allowbreak{}Out\allowbreak{}Of\allowbreak{}Pen\allowbreak{}Holder & 0.0 & 0.0 & 0.0 & \textbf{20.0} & 0.0 & 0.0 \\
\midrule
Avg. & 0.4 & 1.2 & 2.4 & \textbf{37.6} & \underline{34.0} & 17.6 \\
\bottomrule
\end{tabular}
\end{table}

% ==================== Spatial ====================

\begin{table}[t]
\centering
\small
\caption{Per-task in-distribution success rates (\%) on the HUMANO Spatial suite, evaluated over 25 rollouts per task. \reportchange{Within each row, \textbf{bold} and \underline{underline} mark the two highest distinct positive scores, respectively.}}
\label{tab:spatial_tasks}
\setlength{\tabcolsep}{5pt}
\renewcommand{\arraystretch}{1.1}
\vskip -0.05in
\begin{tabular}{@{}>{\raggedright\arraybackslash}p{\dimexpr0.42\linewidth+12pt\relax}>{\centering\arraybackslash}p{40pt}>{\centering\arraybackslash}p{40pt}>{\centering\arraybackslash}p{23pt}>{\centering\arraybackslash}p{23pt}>{\centering\arraybackslash}p{30pt}>{\centering\arraybackslash}p{23pt}@{}}
\toprule
Task & \shortstack{Diffusion\\Policy} & \shortstack{Flow\\Matching} & ACT & $\pi_{0.5}$ & \shortstack{GR00T\\N1.7} & $\Psi_0$ \\
\midrule
\hyperref[tab:tasks_spatial]{\reportchange{S-1}}\reportchange{:}\ Table\allowbreak{}Apple\allowbreak{}Pick & 0.0 & 8.0 & 4.0 & \textbf{36.0} & \underline{28.0} & 8.0 \\
\hyperref[tab:tasks_spatial]{\reportchange{S-2}}\reportchange{:}\ Table\allowbreak{}Plate\allowbreak{}Pick & 0.0 & 4.0 & \underline{8.0} & 4.0 & \textbf{24.0} & 4.0 \\
\hyperref[tab:tasks_spatial]{\reportchange{S-3}}\reportchange{:}\ Ground\allowbreak{}Bottle\allowbreak{}Pick & 0.0 & 0.0 & 0.0 & 0.0 & \textbf{8.0} & 0.0 \\
\hyperref[tab:tasks_spatial]{\reportchange{S-4}}\reportchange{:}\ Ground\allowbreak{}Box\allowbreak{}Pick & 0.0 & 8.0 & 8.0 & \underline{48.0} & \textbf{84.0} & 32.0 \\
\hyperref[tab:tasks_spatial]{\reportchange{S-5}}\reportchange{:}\ Chair\allowbreak{}Sit & 0.0 & 0.0 & 0.0 & \textbf{20.0} & \underline{4.0} & 0.0 \\
\hyperref[tab:tasks_spatial]{\reportchange{S-6}}\reportchange{:}\ Dolly\allowbreak{}Push & 24.0 & 28.0 & 28.0 & \textbf{68.0} & \underline{64.0} & 40.0 \\
\hyperref[tab:tasks_spatial]{\reportchange{S-7}}\reportchange{:}\ Chair\allowbreak{}Push & 0.0 & 4.0 & 0.0 & \underline{40.0} & \textbf{48.0} & 4.0 \\
\hyperref[tab:tasks_spatial]{\reportchange{S-8}}\reportchange{:}\ Ball\allowbreak{}Kick & 20.0 & 16.0 & 28.0 & \underline{52.0} & \textbf{80.0} & 20.0 \\
\hyperref[tab:tasks_spatial]{\reportchange{S-9}}\reportchange{:}\ Table\allowbreak{}Button\allowbreak{}Press & 0.0 & 0.0 & 4.0 & \textbf{44.0} & \underline{20.0} & 0.0 \\
\hyperref[tab:tasks_spatial]{\reportchange{S-10}}\reportchange{:}\ Ground\allowbreak{}Basket\allowbreak{}Pick & 0.0 & 28.0 & 12.0 & 20.0 & \textbf{52.0} & \underline{32.0} \\
\hyperref[tab:tasks_spatial]{\reportchange{S-11}}\reportchange{:}\ Table\allowbreak{}Apple\allowbreak{}To\allowbreak{}Basket & 0.0 & \underline{4.0} & 0.0 & \underline{4.0} & \textbf{16.0} & \underline{4.0} \\
\hyperref[tab:tasks_spatial]{\reportchange{S-12}}\reportchange{:}\ Table\allowbreak{}Plate\allowbreak{}To\allowbreak{}Shelf & 0.0 & 4.0 & 0.0 & \textbf{16.0} & \underline{8.0} & 0.0 \\
\hyperref[tab:tasks_spatial]{\reportchange{S-13}}\reportchange{:}\ Table\allowbreak{}Milk\allowbreak{}To\allowbreak{}Trash\allowbreak{}Can & 0.0 & 0.0 & \textbf{4.0} & 0.0 & 0.0 & \textbf{4.0} \\
\hyperref[tab:tasks_spatial]{\reportchange{S-14}}\reportchange{:}\ Table\allowbreak{}Bread\allowbreak{}To\allowbreak{}Tray & 0.0 & \underline{8.0} & 0.0 & 4.0 & \textbf{36.0} & 0.0 \\
\hyperref[tab:tasks_spatial]{\reportchange{S-15}}\reportchange{:}\ Ground\allowbreak{}Bottle\allowbreak{}To\allowbreak{}Table & 0.0 & \underline{8.0} & 0.0 & 0.0 & \textbf{24.0} & 0.0 \\
\hyperref[tab:tasks_spatial]{\reportchange{S-16}}\reportchange{:}\ Ground\allowbreak{}Box\allowbreak{}To\allowbreak{}Table & 0.0 & 12.0 & 8.0 & 20.0 & \textbf{32.0} & \underline{24.0} \\
\hyperref[tab:tasks_spatial]{\reportchange{S-17}}\reportchange{:}\ Chair\allowbreak{}Push\allowbreak{}To\allowbreak{}Table & 24.0 & \underline{52.0} & \underline{52.0} & \textbf{84.0} & \textbf{84.0} & 48.0 \\
\hyperref[tab:tasks_spatial]{\reportchange{S-18}}\reportchange{:}\ Dolly\allowbreak{}Push\allowbreak{}To\allowbreak{}Table & 24.0 & 12.0 & 36.0 & \underline{64.0} & \textbf{76.0} & 12.0 \\
\hyperref[tab:tasks_spatial]{\reportchange{S-19}}\reportchange{:}\ Table\allowbreak{}Eraser\allowbreak{}Out\allowbreak{}Of\allowbreak{}Pen\allowbreak{}Holder & 0.0 & 0.0 & 0.0 & 0.0 & 0.0 & 0.0 \\
\hyperref[tab:tasks_spatial]{\reportchange{S-20}}\reportchange{:}\ Ground\allowbreak{}Basket\allowbreak{}To\allowbreak{}Table & 0.0 & 0.0 & 4.0 & 16.0 & \textbf{28.0} & \underline{20.0} \\
\midrule
Avg. & 4.6 & 9.8 & 9.8 & \underline{27.0} & \textbf{35.8} & 12.6 \\
\bottomrule
\end{tabular}
\end{table}

% ==================== Object ====================

\begin{table}[t]
\centering
\small
\caption{Per-task in-distribution success rates (\%) on the HUMANO Object suite, evaluated over 25 rollouts per task. \reportchange{Within each row, \textbf{bold} and \underline{underline} mark the two highest distinct positive scores, respectively.}}
\label{tab:object_tasks}
\setlength{\tabcolsep}{5pt}
\renewcommand{\arraystretch}{1.1}
\vskip -0.05in
\begin{tabular}{@{}>{\raggedright\arraybackslash}p{\dimexpr0.42\linewidth+12pt\relax}>{\centering\arraybackslash}p{40pt}>{\centering\arraybackslash}p{40pt}>{\centering\arraybackslash}p{23pt}>{\centering\arraybackslash}p{23pt}>{\centering\arraybackslash}p{30pt}>{\centering\arraybackslash}p{23pt}@{}}
\toprule
Task & \shortstack{Diffusion\\Policy} & \shortstack{Flow\\Matching} & ACT & $\pi_{0.5}$ & \shortstack{GR00T\\N1.7} & $\Psi_0$ \\
\midrule
\hyperref[tab:tasks_object_movement]{\reportchange{O-1}}\reportchange{:}\ \reportchange{Table\allowbreak{}Small\allowbreak{}Object\allowbreak{}Pick} & 28.0 & 24.0 & \underline{36.0} & \textbf{52.0} & \underline{36.0} & 16.0 \\
\hyperref[tab:tasks_object_movement]{\reportchange{O-2}}\reportchange{:}\ \reportchange{Table\allowbreak{}Large\allowbreak{}Object\allowbreak{}Pick} & 32.0 & 40.0 & 52.0 & \underline{80.0} & \textbf{92.0} & 68.0 \\
\hyperref[tab:tasks_object_movement]{\reportchange{O-3}}\reportchange{:}\ \reportchange{Table\allowbreak{}Object\allowbreak{}Pick} & 16.0 & 32.0 & 32.0 & \textbf{44.0} & \textbf{44.0} & \underline{40.0} \\
\hyperref[tab:tasks_object_movement]{\reportchange{O-4}}\reportchange{:}\ \reportchange{Ground\allowbreak{}Small\allowbreak{}Object\allowbreak{}Pick} & 24.0 & 36.0 & \textbf{60.0} & 44.0 & \underline{48.0} & \underline{48.0} \\
\hyperref[tab:tasks_object_movement]{\reportchange{O-5}}\reportchange{:}\ \reportchange{Ground\allowbreak{}Large\allowbreak{}Object\allowbreak{}Pick} & 4.0 & 36.0 & 40.0 & \textbf{56.0} & 40.0 & \underline{52.0} \\
\hyperref[tab:tasks_object_movement]{\reportchange{O-6}}\reportchange{:}\ \reportchange{Ground\allowbreak{}Object\allowbreak{}Pick} & 8.0 & 28.0 & \underline{36.0} & \textbf{40.0} & \textbf{40.0} & 32.0 \\
\hyperref[tab:tasks_object_movement]{\reportchange{O-7}}\reportchange{:}\ \reportchange{Table\allowbreak{}Small\allowbreak{}Object\allowbreak{}To\allowbreak{}Basket} & 12.0 & 36.0 & 40.0 & \underline{48.0} & \textbf{64.0} & 20.0 \\
\hyperref[tab:tasks_object_movement]{\reportchange{O-8}}\reportchange{:}\ \reportchange{Table\allowbreak{}Large\allowbreak{}Object\allowbreak{}To\allowbreak{}Cabinet} & 0.0 & 0.0 & 0.0 & 0.0 & 0.0 & 0.0 \\
\hyperref[tab:tasks_object_movement]{\reportchange{O-9}}\reportchange{:}\ \reportchange{Table\allowbreak{}Object\allowbreak{}To\allowbreak{}Trash\allowbreak{}Can} & 0.0 & 4.0 & 0.0 & 12.0 & \textbf{20.0} & \underline{16.0} \\
\hyperref[tab:tasks_object_movement]{\reportchange{O-10}}\reportchange{:}\ \reportchange{Food\allowbreak{}To\allowbreak{}Tray} & 0.0 & 28.0 & 24.0 & \textbf{40.0} & \underline{32.0} & 24.0 \\
\midrule
Avg. & 12.4 & 26.4 & \underline{32.0} & \textbf{41.6} & \textbf{41.6} & 31.6 \\
\bottomrule
\end{tabular}
\end{table}

\begin{table}[t]
\centering
\small
\caption{Per-task in-distribution success rates (\%) on the HUMANO Movement suite, evaluated over 25 rollouts per task. \reportchange{Within each row, \textbf{bold} and \underline{underline} mark the two highest distinct positive scores, respectively.}}
\label{tab:movement_tasks}
\setlength{\tabcolsep}{5pt}
\renewcommand{\arraystretch}{1.1}
\vskip -0.05in
\begin{tabular}{@{}>{\raggedright\arraybackslash}p{\dimexpr0.42\linewidth+12pt\relax}>{\centering\arraybackslash}p{40pt}>{\centering\arraybackslash}p{40pt}>{\centering\arraybackslash}p{23pt}>{\centering\arraybackslash}p{23pt}>{\centering\arraybackslash}p{30pt}>{\centering\arraybackslash}p{23pt}@{}}
\toprule
Task & \shortstack{Diffusion\\Policy} & \shortstack{Flow\\Matching} & ACT & $\pi_{0.5}$ & \shortstack{GR00T\\N1.7} & $\Psi_0$ \\
\midrule
\hyperref[tab:tasks_object_movement]{\reportchange{M-1}}\reportchange{:}\ Table\allowbreak{}Apple\allowbreak{}Pick & 0.0 & \underline{12.0} & \textbf{16.0} & 0.0 & 4.0 & 4.0 \\
\hyperref[tab:tasks_object_movement]{\reportchange{M-2}}\reportchange{:}\ Table\allowbreak{}Plate\allowbreak{}Pick & 0.0 & 4.0 & 16.0 & 4.0 & \underline{32.0} & \textbf{40.0} \\
\hyperref[tab:tasks_object_movement]{\reportchange{M-3}}\reportchange{:}\ Chair\allowbreak{}Push & 40.0 & 44.0 & 44.0 & 64.0 & \underline{72.0} & \textbf{88.0} \\
\hyperref[tab:tasks_object_movement]{\reportchange{M-4}}\reportchange{:}\ Dolly\allowbreak{}Push & 8.0 & 8.0 & 16.0 & \underline{48.0} & 44.0 & \textbf{68.0} \\
\hyperref[tab:tasks_object_movement]{\reportchange{M-5}}\reportchange{:}\ Chair\allowbreak{}Sit & 0.0 & 0.0 & 0.0 & \textbf{8.0} & 0.0 & 0.0 \\
\hyperref[tab:tasks_object_movement]{\reportchange{M-6}}\reportchange{:}\ Table\allowbreak{}Button\allowbreak{}Press & 0.0 & \textbf{4.0} & \textbf{4.0} & \textbf{4.0} & 0.0 & \textbf{4.0} \\
\hyperref[tab:tasks_object_movement]{\reportchange{M-7}}\reportchange{:}\ Ground\allowbreak{}Bottle\allowbreak{}Pick & 0.0 & \textbf{8.0} & 0.0 & \textbf{8.0} & 0.0 & 0.0 \\
\hyperref[tab:tasks_object_movement]{\reportchange{M-8}}\reportchange{:}\ Ground\allowbreak{}Box\allowbreak{}Pick & 32.0 & 16.0 & 56.0 & \textbf{96.0} & 76.0 & \underline{80.0} \\
\hyperref[tab:tasks_object_movement]{\reportchange{M-9}}\reportchange{:}\ Ball\allowbreak{}Kick & 28.0 & 76.0 & 76.0 & \underline{84.0} & \textbf{92.0} & 80.0 \\
\hyperref[tab:tasks_object_movement]{\reportchange{M-10}}\reportchange{:}\ Ground\allowbreak{}Basket\allowbreak{}Pick & 8.0 & 28.0 & \textbf{68.0} & \textbf{68.0} & \underline{44.0} & 20.0 \\
\midrule
Avg. & 11.6 & 20.0 & 29.6 & \textbf{38.4} & \underline{36.4} & \textbf{38.4} \\
\bottomrule
\end{tabular}
\end{table}

% ==================== Goal ====================
\begin{table}[t]
\centering
\small
\caption{Per-task in-distribution success rates (\%) on the HUMANO Goal suite, evaluated over 25 rollouts per task. \reportchange{Within each row, \textbf{bold} and \underline{underline} mark the two highest distinct positive scores, respectively.}}
\label{tab:goal_tasks}
\setlength{\tabcolsep}{5pt}
\renewcommand{\arraystretch}{1.1}
\vskip -0.05in
\begin{tabular}{@{}>{\raggedright\arraybackslash}p{\dimexpr0.42\linewidth+12pt\relax}>{\centering\arraybackslash}p{40pt}>{\centering\arraybackslash}p{40pt}>{\centering\arraybackslash}p{23pt}>{\centering\arraybackslash}p{23pt}>{\centering\arraybackslash}p{30pt}>{\centering\arraybackslash}p{23pt}@{}}
\toprule
Task & \shortstack{Diffusion\\Policy} & \shortstack{Flow\\Matching} & ACT & $\pi_{0.5}$ & \shortstack{GR00T\\N1.7} & $\Psi_0$ \\
\midrule
\hyperref[tab:tasks_goal]{\reportchange{G-1}}\reportchange{:}\ Goal\allowbreak{}Apple\allowbreak{}Placement & 0.0 & 12.0 & 12.0 & \underline{16.0} & \textbf{36.0} & 0.0 \\
\hyperref[tab:tasks_goal]{\reportchange{G-2}}\reportchange{:}\ Goal\allowbreak{}Plate\allowbreak{}Placement & 0.0 & 4.0 & 8.0 & \underline{16.0} & \textbf{32.0} & \underline{16.0} \\
\hyperref[tab:tasks_goal]{\reportchange{G-3}}\reportchange{:}\ Goal\allowbreak{}Bread\allowbreak{}Placement & 0.0 & 4.0 & 0.0 & \underline{16.0} & \textbf{40.0} & 0.0 \\
\hyperref[tab:tasks_goal]{\reportchange{G-4}}\reportchange{:}\ Goal\allowbreak{}Milk\allowbreak{}Placement & 0.0 & \textbf{16.0} & 0.0 & \underline{8.0} & \underline{8.0} & 0.0 \\
\hyperref[tab:tasks_goal]{\reportchange{G-5}}\reportchange{:}\ Goal\allowbreak{}Object\allowbreak{}To\allowbreak{}Basket & 0.0 & 0.0 & \underline{4.0} & 0.0 & \textbf{24.0} & 0.0 \\
\hyperref[tab:tasks_goal]{\reportchange{G-6}}\reportchange{:}\ Goal\allowbreak{}Book\allowbreak{}Placement & 0.0 & 0.0 & 0.0 & \textbf{12.0} & \underline{8.0} & \underline{8.0} \\
\hyperref[tab:tasks_goal]{\reportchange{G-7}}\reportchange{:}\ Goal\allowbreak{}Ground\allowbreak{}Bottle\allowbreak{}Placement & 0.0 & 0.0 & 0.0 & 0.0 & \textbf{8.0} & 0.0 \\
\hyperref[tab:tasks_goal]{\reportchange{G-8}}\reportchange{:}\ Goal\allowbreak{}Eraser\allowbreak{}Pour & 0.0 & 8.0 & \underline{12.0} & \underline{12.0} & \textbf{20.0} & 4.0 \\
\hyperref[tab:tasks_goal]{\reportchange{G-9}}\reportchange{:}\ Goal\allowbreak{}Storage\allowbreak{}Box\allowbreak{}Placement & 0.0 & 0.0 & 0.0 & 0.0 & 0.0 & 0.0 \\
\hyperref[tab:tasks_goal]{\reportchange{G-10}}\reportchange{:}\ Goal\allowbreak{}Football\allowbreak{}Placement & \textbf{36.0} & 0.0 & 0.0 & 8.0 & 16.0 & \underline{20.0} \\
\midrule
Avg. & 3.6 & 4.4 & 3.6 & \underline{8.8} & \textbf{19.2} & 4.8 \\
\bottomrule
\end{tabular}
\end{table}

% ==================== Long ====================
\begin{table*}[t]
\centering
\small
\caption{Per-task in-distribution success rates (\%) on the HUMANO Long suite, evaluated over 25 rollouts per task. \reportchange{Within each row, \textbf{bold} and \underline{underline} mark the two highest distinct positive scores, respectively.}}
\label{tab:long_tasks}
\setlength{\tabcolsep}{5pt}
\renewcommand{\arraystretch}{1.1}
\vskip -0.05in
\begin{tabular}{@{}>{\raggedright\arraybackslash}p{\dimexpr0.42\linewidth+12pt\relax}>{\centering\arraybackslash}p{40pt}>{\centering\arraybackslash}p{40pt}>{\centering\arraybackslash}p{23pt}>{\centering\arraybackslash}p{23pt}>{\centering\arraybackslash}p{30pt}>{\centering\arraybackslash}p{23pt}@{}}
\toprule
Task & \shortstack{Diffusion\\Policy} & \shortstack{Flow\\Matching} & ACT & $\pi_{0.5}$ & \shortstack{GR00T\\N1.7} & $\Psi_0$ \\
\midrule
\hyperref[tab:tasks_long]{\reportchange{L-1}}\reportchange{:}\ Long\allowbreak{}Table\allowbreak{}Apple\allowbreak{}To\allowbreak{}Drawer & 0.0 & 0.0 & 0.0 & 0.0 & 0.0 & 0.0 \\
\hyperref[tab:tasks_long]{\reportchange{L-2}}\reportchange{:}\ Long\allowbreak{}Table\allowbreak{}Plate\allowbreak{}To\allowbreak{}Drawer & 0.0 & 0.0 & 0.0 & 0.0 & \textbf{4.0} & 0.0 \\
\hyperref[tab:tasks_long]{\reportchange{L-3}}\reportchange{:}\ Long\allowbreak{}Bread\allowbreak{}Cake\allowbreak{}To\allowbreak{}Tray & 0.0 & \underline{8.0} & 4.0 & 4.0 & \textbf{12.0} & 0.0 \\
\hyperref[tab:tasks_long]{\reportchange{L-4}}\reportchange{:}\ Long\allowbreak{}Bread\allowbreak{}To\allowbreak{}Tray\allowbreak{}And\allowbreak{}Lift & 0.0 & \underline{4.0} & 0.0 & \underline{4.0} & \textbf{44.0} & 0.0 \\
\hyperref[tab:tasks_long]{\reportchange{L-5}}\reportchange{:}\ Long\allowbreak{}Cake\allowbreak{}To\allowbreak{}Plate\allowbreak{}And\allowbreak{}Lift & 0.0 & 0.0 & 0.0 & 0.0 & 0.0 & 0.0 \\
\hyperref[tab:tasks_long]{\reportchange{L-6}}\reportchange{:}\ Long\allowbreak{}Ground\allowbreak{}Box\allowbreak{}To\allowbreak{}Dolly\allowbreak{}And\allowbreak{}Push & 0.0 & \underline{16.0} & 0.0 & 12.0 & \textbf{28.0} & 12.0 \\
\hyperref[tab:tasks_long]{\reportchange{L-7}}\reportchange{:}\ Long\allowbreak{}Chair\allowbreak{}Push\allowbreak{}And\allowbreak{}Sit & 0.0 & 0.0 & \underline{4.0} & \underline{4.0} & \textbf{8.0} & 0.0 \\
\hyperref[tab:tasks_long]{\reportchange{L-8}}\reportchange{:}\ Long\allowbreak{}Table\allowbreak{}Football\allowbreak{}To\allowbreak{}Ground\allowbreak{}And\allowbreak{}Kick & 12.0 & 36.0 & 16.0 & \underline{44.0} & \textbf{68.0} & 8.0 \\
\hyperref[tab:tasks_long]{\reportchange{L-9}}\reportchange{:}\ Long\allowbreak{}Ground\allowbreak{}Basket\allowbreak{}To\allowbreak{}Table\allowbreak{}Apple\allowbreak{}To\allowbreak{}Basket & 0.0 & 0.0 & 0.0 & 0.0 & \textbf{12.0} & 0.0 \\
\hyperref[tab:tasks_long]{\reportchange{L-10}}\reportchange{:}\ \reportchange{Long\allowbreak{}Ground\allowbreak{}Bottle\allowbreak{}To\allowbreak{}Box\allowbreak{}And\allowbreak{}Lift} & 0.0 & 0.0 & 0.0 & \underline{8.0} & \textbf{20.0} & 0.0 \\
\hyperref[tab:tasks_long]{\reportchange{L-11}}\reportchange{:}\ Long\allowbreak{}Apple\allowbreak{}Banana\allowbreak{}To\allowbreak{}Basket & 0.0 & 0.0 & 0.0 & 0.0 & \textbf{4.0} & 0.0 \\
\hyperref[tab:tasks_long]{\reportchange{L-12}}\reportchange{:}\ Long\allowbreak{}Basket\allowbreak{}Apple\allowbreak{}To\allowbreak{}Table\allowbreak{}Banana\allowbreak{}To\allowbreak{}Trash\allowbreak{}Can & 0.0 & 0.0 & 0.0 & 0.0 & \textbf{12.0} & 0.0 \\
\hyperref[tab:tasks_long]{\reportchange{L-13}}\reportchange{:}\ Long\allowbreak{}Peanuts\allowbreak{}From\allowbreak{}Cup\allowbreak{}To\allowbreak{}Plate & 0.0 & 0.0 & 0.0 & 0.0 & \textbf{4.0} & 0.0 \\
\hyperref[tab:tasks_long]{\reportchange{L-14}}\reportchange{:}\ Long\allowbreak{}Storage\allowbreak{}Box\allowbreak{}Cake\allowbreak{}To\allowbreak{}Table\allowbreak{}Bread\allowbreak{}To\allowbreak{}Tray & 0.0 & 0.0 & 0.0 & 0.0 & \textbf{4.0} & 0.0 \\
\hyperref[tab:tasks_long]{\reportchange{L-15}}\reportchange{:}\ Long\allowbreak{}Apple\allowbreak{}To\allowbreak{}Basket\allowbreak{}Then\allowbreak{}Basket\allowbreak{}To\allowbreak{}Ground & 0.0 & 0.0 & 0.0 & 0.0 & \textbf{32.0} & 0.0 \\
\hyperref[tab:tasks_long]{\reportchange{L-16}}\reportchange{:}\ Long\allowbreak{}High\allowbreak{}Table\allowbreak{}Apple\allowbreak{}To\allowbreak{}Low\allowbreak{}Table\allowbreak{}Drawer & 0.0 & 0.0 & 0.0 & 0.0 & 0.0 & 0.0 \\
\hyperref[tab:tasks_long]{\reportchange{L-17}}\reportchange{:}\ \reportchange{Long\allowbreak{}Low\allowbreak{}Table\allowbreak{}Plate\allowbreak{}To\allowbreak{}High\allowbreak{}Table\allowbreak{}Drawer} & \reportchange{0.0} & \reportchange{0.0} & \reportchange{0.0} & \reportchange{0.0} & \reportchange{\textbf{4.0}} & \reportchange{0.0} \\
\hyperref[tab:tasks_long]{\reportchange{L-18}}\reportchange{:}\ Long\allowbreak{}Move\allowbreak{}Box\allowbreak{}Then\allowbreak{}Chair\allowbreak{}To\allowbreak{}Table & 0.0 & 0.0 & 0.0 & 0.0 & \textbf{20.0} & 0.0 \\
\hyperref[tab:tasks_long]{\reportchange{L-19}}\reportchange{:}\ Table\allowbreak{}Book\allowbreak{}To\allowbreak{}Shelf & 0.0 & 0.0 & 0.0 & 0.0 & \textbf{16.0} & 0.0 \\
\hyperref[tab:tasks_long]{\reportchange{L-20}}\reportchange{:}\ Long\allowbreak{}Shelf\allowbreak{}Book\allowbreak{}To\allowbreak{}Table\allowbreak{}Drawer & 0.0 & 0.0 & 0.0 & 0.0 & 0.0 & 0.0 \\
\midrule
Avg. & 0.6 & \reportchange{3.2} & \reportchange{1.2} & \reportchange{\underline{3.8}} & \reportchange{\textbf{14.6}} & \reportchange{1.0} \\
\bottomrule
\end{tabular}
\end{table*}

% ==================== Conditional ====================
\begin{table}[t]
\centering
\small
\caption{Per-task in-distribution success rates (\%) on the HUMANO Conditional suite, evaluated over 25 rollouts per task. \reportchange{Within each row, \textbf{bold} and \underline{underline} mark the two highest distinct positive scores, respectively.}}
\label{tab:conditional_tasks}
\setlength{\tabcolsep}{5pt}
\renewcommand{\arraystretch}{1.1}
\vskip -0.05in
\begin{tabular}{@{}>{\raggedright\arraybackslash}p{\dimexpr0.42\linewidth+12pt\relax}>{\centering\arraybackslash}p{40pt}>{\centering\arraybackslash}p{40pt}>{\centering\arraybackslash}p{23pt}>{\centering\arraybackslash}p{23pt}>{\centering\arraybackslash}p{30pt}>{\centering\arraybackslash}p{23pt}@{}}
\toprule
Task & \shortstack{Diffusion\\Policy} & \shortstack{Flow\\Matching} & ACT & $\pi_{0.5}$ & \shortstack{GR00T\\N1.7} & $\Psi_0$ \\
\midrule
\hyperref[tab:tasks_conditional]{\reportchange{C-1}}\reportchange{:}\ Condition\allowbreak{}Plate\allowbreak{}To\allowbreak{}Drawer & 0.0 & \textbf{8.0} & 0.0 & 0.0 & \textbf{8.0} & 0.0 \\
\hyperref[tab:tasks_conditional]{\reportchange{C-2}}\reportchange{:}\ Condition\allowbreak{}Apple\allowbreak{}To\allowbreak{}Drawer & 0.0 & \underline{12.0} & 0.0 & 4.0 & \textbf{20.0} & 0.0 \\
\hyperref[tab:tasks_conditional]{\reportchange{C-3}}\reportchange{:}\ Condition\allowbreak{}Bottle\allowbreak{}To\allowbreak{}Empty\allowbreak{}Box & 0.0 & \underline{4.0} & 0.0 & \textbf{8.0} & \underline{4.0} & 0.0 \\
\hyperref[tab:tasks_conditional]{\reportchange{C-4}}\reportchange{:}\ Condition\allowbreak{}Apple\allowbreak{}To\allowbreak{}Fruit\allowbreak{}Free\allowbreak{}Basket & 4.0 & 4.0 & 8.0 & \underline{16.0} & \textbf{44.0} & 4.0 \\
\hyperref[tab:tasks_conditional]{\reportchange{C-5}}\reportchange{:}\ Condition\allowbreak{}Chair\allowbreak{}To\allowbreak{}Table\allowbreak{}And\allowbreak{}Sit & 0.0 & 0.0 & 0.0 & \textbf{20.0} & \underline{8.0} & 0.0 \\
\hyperref[tab:tasks_conditional]{\reportchange{C-6}}\reportchange{:}\ Condition\allowbreak{}Basket\allowbreak{}To\allowbreak{}Table\allowbreak{}Apple\allowbreak{}To\allowbreak{}Basket & 0.0 & 12.0 & 16.0 & \textbf{36.0} & \underline{24.0} & 20.0 \\
\hyperref[tab:tasks_conditional]{\reportchange{C-7}}\reportchange{:}\ Condition\allowbreak{}Peanuts\allowbreak{}To\allowbreak{}Empty\allowbreak{}Plate\allowbreak{}Or\allowbreak{}Tray & 0.0 & 8.0 & 8.0 & \underline{20.0} & \textbf{48.0} & 0.0 \\
\hyperref[tab:tasks_conditional]{\reportchange{C-8}}\reportchange{:}\ Condition\allowbreak{}Bread\allowbreak{}To\allowbreak{}Plate\allowbreak{}Or\allowbreak{}Tray & 0.0 & 0.0 & \underline{4.0} & \underline{4.0} & \textbf{40.0} & \underline{4.0} \\
\hyperref[tab:tasks_conditional]{\reportchange{C-9}}\reportchange{:}\ \reportchange{Condition\allowbreak{}Clear\allowbreak{}Dolly\allowbreak{}And\allowbreak{}Push\allowbreak{}Chair} & 0.0 & 16.0 & \underline{36.0} & 24.0 & \textbf{44.0} & \underline{36.0} \\
\hyperref[tab:tasks_conditional]{\reportchange{C-10}}\reportchange{:}\ Condition\allowbreak{}Book\allowbreak{}To\allowbreak{}Occupied\allowbreak{}Shelf\allowbreak{}Layer & 0.0 & \underline{4.0} & 0.0 & \underline{4.0} & \textbf{8.0} & 0.0 \\
\midrule
Avg. & 0.4 & 6.8 & 7.2 & \underline{13.6} & \textbf{24.8} & 6.4 \\
\bottomrule
\end{tabular}
\end{table}

\subsection{Evaluation Protocol}
\label{app:baseline_evaluation}

All six baselines are evaluated using the same simulation and control pipeline. MuJoCo simulates robot and object dynamics, while Isaac Sim renders the corresponding visual observations. Each episode is initialized from a saved evaluation configuration containing the robot state, task objects, and scene configuration. Policy states and action buffers are reset between episodes, and language-conditioned policies receive the corresponding task instruction.

Each predicted 78-dimensional action is decomposed into a 64-dimensional SONIC motion token and two 7-dimensional hand commands. The SONIC controller converts the motion token into whole-body motor commands, while the hand commands control the corresponding hand joints. Predicted action chunks are stored in an execution queue and consumed sequentially at the action rate defined by the dataset. When the queue is exhausted, the policy is queried again using the latest observation. The low-level controller continues executing the current motion command between policy updates.

The rollout horizon depends on the task suite. Long-Horizon tasks are allowed at most 7,000 environment steps, while tasks in the other seven suites are limited to 3,000:
\begin{equation}
T_{\max}^{(s)} =
\begin{cases}
7{,}000, & s=\mathrm{Long\text{-}Horizon},\\
3{,}000, & \text{otherwise}.
\end{cases}
\end{equation}
These limits count environment interactions during policy execution and exclude robot initialization, stabilization, controller warm-up, and action prefetching. An episode terminates when the task-specific success condition is satisfied, the environment terminates or truncates, the step budget is exhausted, or an execution-ending failure occurs. Episodes that reach the maximum horizon without completing the task are counted as failures.

Task completion is determined by the task-specific success checker implemented in the environment. The final outcome additionally accounts for robot falls and detected scene-wall collisions. A fall is detected when a core body part---the pelvis, waist, or torso---remains in contact with the ground for at least 0.5\,s. An episode is counted as successful only when the task success condition is satisfied without a recorded fall or execution-ending collision.

The primary evaluation metric is task success rate. For a suite containing $K$ tasks, we compute
\begin{equation}
\mathrm{SR}_{\mathrm{suite}}
=
\frac{1}{K}
\sum_{k=1}^{K}
\left(
\frac{1}{N_k}
\sum_{i=1}^{N_k} y_{k,i}
\right),
\end{equation}
where $N_k$ is the number of evaluation episodes for task $k$ and $y_{k,i}\in\{0,1\}$ denotes the final outcome of episode $i$. Under the standard HUMANO protocol, all tasks within the same evaluation setting use the same number of rollouts, making this equivalent to pooling episode outcomes within the suite.

All compared policies use the same evaluation configurations, execution interface, task success criteria, and suite-specific rollout budgets. Model configurations, preprocessing statistics, and training states required for inference are retained in the corresponding checkpoints. The HUMANO-specific modifications are limited to dataset conversion, observation and action interfaces, embodiment registration, checkpoint compatibility, and launcher configuration; the core architectures and learning objectives remain those of the original baseline implementations.

\section{Implementation Details of Ablation and Analysis Study}
\label{app:ablation_analysis}

\subsection{Implementation Details of Pretrain}
\label{app:pretrain_details}

\paragraph{Training Data and Strategy.} We adopt a two-stage training strategy initialized from the pretrained GR00T-N1.7-3B checkpoint. In the first stage, we perform continued pretraining for 10 effective epochs using either HUMANO or SIMPLE G1. \reportchange{HUMANO pretraining used 5K demonstrations across eight suites. SIMPLE G1 pretraining used about 1.3K demonstrations across 14 tasks.} In the second stage, both models undergo 5 effective epochs of post-training on the same HUMANO Basic dataset, which contains 10 tasks and 500 demonstrations. The HUMANO and SIMPLE pretraining stages involve 77,980 and 7,690 optimization steps, respectively, followed by 2,105 steps of post-training.

\paragraph{Optimization Details.} Both stages are conducted on 8 GPUs with a per-GPU batch size of 64, resulting in a global batch size of 512. We employ the AdamW optimizer with a learning rate of $2\times10^{-4}$, weight decay of $10^{-5}$, cosine learning rate scheduling, and a warmup ratio of 0.05. Training uses BF16 mixed precision, gradient clipping at 1.0, and a state dropout rate of 0.2. We freeze the language model and visual backbone while updating the projector, diffusion model, and VLLN modules. Checkpoints are saved after each effective epoch.

\paragraph{Data Processing and Model Adaptation.} Input RGB images are resized from $640\times360$ to $256\times256$ using letterbox padding, random cropping, and color jittering during training. The action window lengths are 40 for HUMANO and 30 for SIMPLE. HUMANO uses 46-dimensional states and 78-dimensional actions, whereas SIMPLE uses 32-dimensional states and 36-dimensional actions. For SIMPLE-to-Basic transfer, we expand the state/action dimensions to 46/78 by preserving overlapping pretrained weights and initializing newly introduced dimensions. For HUMANO-to-Basic, no dimensional expansion is required.

\section{Additional Experiments}
\label{app:additional_experiments}

\subsection{Task-wise Results of In-distribution Setting}

As shown in Tables~\ref{tab:basic_tasks}--\ref{tab:conditional_tasks}, task-level performance varies both within and across suites. In the Basic suite, the strongest policies achieve relatively high success on \emph{DollyPush}, whereas tasks such as \emph{ChairSit} and \emph{GroundBottlePick} remain particularly challenging. Similar variation is observed in the Advanced, Spatial, and Object suites, where high success on individual tasks coexists with complete failure on others. \newer{The Movement suite exhibits relatively strong performance on tasks such as \emph{GroundBoxPick} and \emph{BallKick}, while the Goal and Long suites remain particularly challenging, with several tasks unsolved by all evaluated policies. In the Conditional suite, success rates remain limited across most tasks, reflecting the difficulty of state-dependent decision making.} Moreover, the best-performing policy varies across tasks, indicating complementary rather than uniform capabilities across generalist models. Overall, these results show that suite-level averages can obscure substantial task-specific differences and motivate fine-grained evaluation across diverse humanoid behaviors.

\begin{table*}[t]
\centering
\small
\caption{
Out-of-distribution generalization on the \emph{Basic} suite.
We report success rate (\%) under seven controlled OOD settings, averaged over 25 rollouts per task.
\reportchange{Within each column, \textbf{bold} and \underline{underline} mark the two highest distinct positive scores, respectively.}
}
\label{tab:ood_basic}
\vskip -0.05in
\resizebox{\textwidth}{!}{
\begin{tabular}{lcccccccc}
\toprule
Method & Distractor & Lighting & Background & Camera & Language & Sensor & Robot Init. & OOD Avg. \\
\midrule
Diffusion Policy & 2.8 & 0.4 & 5.6 & 0.0 & 2.0 & 0.0 & 0.4 & 1.6 \\
Flow Matching & 2.4 & 0.8 & 2.4 & 0.8 & 4.8 & 1.6 & 1.6 & 2.1 \\
ACT & 1.2 & 1.2 & 8.0 & 0.0 & 5.2 & 0.4 & 3.2 & 2.7 \\
$\pi_{0.5}$ & \textbf{17.6} & \underline{15.2} & \underline{15.6} & \textbf{16.0} & \textbf{22.8} & \textbf{11.2} & \textbf{14.0} & \textbf{16.1} \\
GR00T N1.7 & \underline{17.2} & \textbf{16.0} & \textbf{24.8} & \underline{8.8} & \underline{22.0} & \underline{9.2} & \underline{12.0} & \underline{15.7} \\
$\Psi_0$ & 4.0 & 6.0 & 8.8 & 4.8 & 17.2 & 4.0 & 4.0 & 7.0 \\
\bottomrule
\end{tabular}
}
\end{table*}
\begin{table*}[t]
\centering
\small
\caption{
Out-of-distribution generalization on the \emph{Advanced} suite.
We report success rate (\%) under seven controlled OOD settings, averaged over 25 rollouts per task.
\reportchange{Within each column, \textbf{bold} and \underline{underline} mark the two highest distinct positive scores, respectively.}
}
\label{tab:ood_advanced}
\vskip -0.05in
\resizebox{\textwidth}{!}{
\begin{tabular}{lcccccccc}
\toprule
Method & Distractor & Lighting & Background & Camera & Language & Sensor & Robot Init. & OOD Avg. \\
\midrule
Diffusion Policy & \newer{0.0} & \newer{0.4} & \newer{0.0} & \newer{0.0} & \newer{0.0} & \newer{0.0} & \newer{0.0} & \newer{0.1} \\
Flow Matching & \newer{1.6} & \newer{0.8} & \newer{0.8} & \newer{0.0} & \newer{2.4} & \newer{0.0} & \newer{0.8} & \newer{0.9} \\
ACT & \newer{1.6} & \newer{0.8} & \newer{0.4} & \newer{0.0} & \newer{1.6} & \newer{0.0} & \newer{0.0} & \newer{0.6} \\
$\pi_{0.5}$ & \newer{\textbf{32.4}} & \newer{\textbf{30.4}} & \newer{\underline{29.2}} & \newer{\textbf{20.8}} & \newer{\textbf{36.4}} & \newer{2.0} & \newer{\textbf{20.8}} & \newer{\textbf{24.6}} \\
GR00T N1.7 & \newer{\underline{19.2}} & \newer{\underline{30.0}} & \newer{\textbf{33.2}} & \newer{8.8} & \newer{\underline{28.0}} & \newer{\textbf{8.8}} & \newer{\underline{18.0}} & \newer{\underline{20.9}} \\
$\Psi_0$ & \newer{11.6} & \newer{16.0} & \newer{12.8} & \newer{\underline{14.0}} & \newer{9.2} & \newer{\underline{2.4}} & \newer{12.0} & \newer{11.1} \\
\bottomrule
\end{tabular}
}
\end{table*}
\begin{table*}[!t]
\centering
\small
\caption{
Out-of-distribution generalization on the \emph{Spatial} suite.
We report success rate (\%) under seven controlled OOD settings, averaged over 25 rollouts per task.
\reportchange{Within each column, \textbf{bold} and \underline{underline} mark the two highest distinct positive scores, respectively.}
}
\label{tab:ood_spatial}
\vskip -0.05in
\resizebox{\textwidth}{!}{
\begin{tabular}{lcccccccc}
\toprule
Method & Distractor & Lighting & Background & Camera & Language & Sensor & Robot Init. & OOD Avg. \\
\midrule
Diffusion Policy & \newer{2.6} & \newer{0.4} & \newer{5.0} & \newer{0.0} & \newer{4.2} & \newer{0.2} & \newer{3.8} & \newer{2.3} \\
Flow Matching & \newer{7.0} & \newer{1.4} & \newer{9.8} & \newer{1.0} & \newer{7.2} & \newer{\underline{3.4}} & \newer{11.2} & \newer{5.9} \\
ACT & \newer{6.8} & \newer{0.6} & \newer{8.8} & \newer{0.2} & \newer{10.6} & \newer{0.0} & \newer{9.8} & \newer{5.3} \\
$\pi_{0.5}$ & \newer{\underline{17.6}} & \newer{\underline{18.2}} & \newer{\underline{23.4}} & \newer{\underline{11.6}} & \newer{\underline{21.8}} & \newer{3.2} & \newer{\underline{20.0}} & \newer{\underline{16.5}} \\
GR00T N1.7 & \newer{\textbf{21.0}} & \newer{\textbf{29.4}} & \newer{\textbf{34.6}} & \newer{\textbf{12.0}} & \newer{\textbf{32.4}} & \newer{\textbf{14.6}} & \newer{\textbf{24.6}} & \newer{\textbf{24.1}} \\
$\Psi_0$ & \newer{11.0} & \newer{9.0} & \newer{13.0} & \newer{3.2} & \newer{12.6} & \newer{0.4} & \newer{9.6} & \newer{8.4} \\
\bottomrule
\end{tabular}
}
\end{table*}
\begin{table*}[!t]
\centering
\small
\caption{
Out-of-distribution generalization on the \emph{Object} suite.
We report success rate (\%) under seven controlled OOD settings, averaged over 25 rollouts per task.
\reportchange{Within each column, \textbf{bold} and \underline{underline} mark the two highest distinct positive scores, respectively.}
}
\label{tab:ood_object}
\vskip -0.05in
\resizebox{\textwidth}{!}{
\begin{tabular}{lcccccccc}
\toprule
Method & Distractor & Lighting & Background & Camera & Language & Sensor & Robot Init. & OOD Avg. \\
\midrule
Diffusion Policy & \newer{11.2} & \newer{4.0} & \newer{12.0} & \newer{0.4} & \newer{11.6} & \newer{0.8} & \newer{2.8} & \newer{6.1} \\
Flow Matching & \newer{13.6} & \newer{10.4} & \newer{26.0} & \newer{6.4} & \newer{24.0} & \newer{4.0} & \newer{18.4} & \newer{14.7} \\
ACT & \newer{8.0} & \newer{8.4} & \newer{26.4} & \newer{4.0} & \newer{32.8} & \newer{1.2} & \newer{20.0} & \newer{14.4} \\
$\pi_{0.5}$ & \newer{\underline{38.4}} & \newer{\underline{35.6}} & \newer{\textbf{41.2}} & \newer{\textbf{25.6}} & \newer{\textbf{42.8}} & \newer{\underline{16.8}} & \newer{\textbf{26.4}} & \newer{\textbf{32.4}} \\
GR00T N1.7 & \newer{\textbf{39.6}} & \newer{\textbf{39.6}} & \newer{\underline{38.8}} & \newer{\underline{21.6}} & \newer{\underline{38.8}} & \newer{\textbf{25.2}} & \newer{18.4} & \newer{\underline{31.7}} \\
$\Psi_0$ & \newer{19.2} & \newer{28.0} & \newer{29.2} & \newer{7.2} & \newer{29.2} & \newer{2.4} & \newer{\underline{20.8}} & \newer{19.4} \\
\bottomrule
\end{tabular}
}
\end{table*}
\begin{table*}[t]
\centering
\small
\caption{
Out-of-distribution generalization on the \emph{Movement} suite.
We report success rate (\%) under seven controlled OOD settings, averaged over 25 rollouts per task.
\reportchange{Within each column, \textbf{bold} and \underline{underline} mark the two highest distinct positive scores, respectively.}
}
\label{tab:ood_movement}
\vskip -0.05in
\resizebox{\textwidth}{!}{
\begin{tabular}{lcccccccc}
\toprule
Method & Distractor & Lighting & Background & Camera & Language & Sensor & Robot Init. & OOD Avg. \\
\midrule
Diffusion Policy & \newer{10.0} & \newer{5.2} & \newer{12.8} & \newer{0.0} & \newer{12.0} & \newer{0.0} & \newer{8.8} & \newer{7.0} \\
Flow Matching & \newer{11.6} & \newer{11.6} & \newer{23.2} & \newer{8.4} & \newer{20.4} & \newer{2.8} & \newer{18.0} & \newer{13.7} \\
ACT & \newer{15.6} & \newer{8.4} & \newer{22.8} & \newer{1.6} & \newer{30.4} & \newer{0.0} & \newer{21.6} & \newer{14.3} \\
$\pi_{0.5}$ & \newer{\textbf{37.2}} & \newer{\underline{38.0}} & \newer{\textbf{42.0}} & \newer{\textbf{34.0}} & \newer{\textbf{38.8}} & \newer{\textbf{25.2}} & \newer{\underline{26.0}} & \newer{\textbf{34.5}} \\
GR00T N1.7 & \newer{\underline{30.4}} & \newer{\textbf{40.0}} & \newer{\underline{36.8}} & \newer{\underline{26.4}} & \newer{\underline{37.6}} & \newer{\underline{8.0}} & \newer{\textbf{29.2}} & \newer{\underline{29.8}} \\
$\Psi_0$ & \newer{25.2} & \newer{31.6} & \newer{27.2} & \newer{21.2} & \newer{30.0} & \newer{4.8} & \newer{19.2} & \newer{22.7} \\
\bottomrule
\end{tabular}
}
\end{table*}
\begin{table*}[!t]
\centering
\small
\caption{
Out-of-distribution generalization on the \emph{Goal} suite.
We report success rate (\%) under seven controlled OOD settings, averaged over 25 rollouts per task.
\reportchange{Within each column, \textbf{bold} and \underline{underline} mark the two highest distinct positive scores, respectively.}
}
\label{tab:ood_goal}
\vskip -0.05in
\resizebox{\textwidth}{!}{
\begin{tabular}{lcccccccc}
\toprule
Method & Distractor & Lighting & Background & Camera & Language & Sensor & Robot Init. & OOD Avg. \\
\midrule
Diffusion Policy & \newer{2.0} & \newer{2.8} & \newer{1.2} & \newer{0.4} & \newer{2.4} & \newer{\underline{1.6}} & \newer{0.8} & \newer{1.6} \\
Flow Matching & \newer{4.4} & \newer{0.8} & \newer{4.4} & \newer{1.2} & \newer{4.8} & \newer{0.4} & \newer{3.2} & \newer{2.7} \\
ACT & \newer{1.2} & \newer{0.0} & \newer{6.8} & \newer{0.4} & \newer{2.4} & \newer{1.2} & \newer{0.8} & \newer{1.8} \\
$\pi_{0.5}$ & \newer{\underline{10.0}} & \newer{\underline{7.2}} & \newer{\underline{10.0}} & \newer{\underline{1.6}} & \newer{\underline{6.8}} & \newer{1.2} & \newer{\textbf{8.4}} & \newer{\underline{6.5}} \\
GR00T N1.7 & \newer{\textbf{15.2}} & \newer{\textbf{14.4}} & \newer{\textbf{20.8}} & \newer{\textbf{2.0}} & \newer{\textbf{19.2}} & \newer{\textbf{2.0}} & \newer{\textbf{8.4}} & \newer{\textbf{11.7}} \\
$\Psi_0$ & \newer{4.0} & \newer{1.6} & \newer{3.2} & \newer{0.4} & \newer{3.2} & \newer{0.8} & \newer{\underline{4.0}} & \newer{2.5} \\
\bottomrule
\end{tabular}
}
\end{table*}
\begin{table*}[!t]
\centering
\small
\caption{
Out-of-distribution generalization on the \emph{Long} suite.
We report success rate (\%) under seven controlled OOD settings, averaged over 25 rollouts per task.
\reportchange{Within each column, \textbf{bold} and \underline{underline} mark the two highest distinct positive scores, respectively.}
}
\label{tab:ood_long}
\vskip -0.05in
\resizebox{\textwidth}{!}{
\begin{tabular}{lcccccccc}
\toprule
Method & Distractor & Lighting & Background & Camera & Language & Sensor & Robot Init. & OOD Avg. \\
\midrule
Diffusion Policy & \newer{0.2} & \newer{0.0} & \newer{0.0} & \newer{0.0} & \newer{0.4} & \newer{0.0} & \newer{0.4} & \newer{0.1} \\
Flow Matching & \reportchange{1.6} & \reportchange{0.6} & \reportchange{2.6} & \reportchange{0.2} & \reportchange{\underline{4.4}} & \reportchange{0.0} & \reportchange{2.6} & \reportchange{1.7} \\
ACT & \newer{0.8} & \newer{0.0} & \newer{1.2} & \newer{0.2} & \newer{1.4} & \newer{\underline{1.6}} & \newer{1.4} & \newer{0.9} \\
$\pi_{0.5}$ & \newer{\underline{4.6}} & \newer{\underline{4.4}} & \newer{\underline{4.0}} & \reportchange{\textbf{1.8}} & \newer{\underline{4.4}} & \newer{\textbf{1.8}} & \newer{\underline{4.6}} & \newer{\underline{3.7}} \\
GR00T N1.7 & \newer{\textbf{11.2}} & \newer{\textbf{10.4}} & \newer{\textbf{16.8}} & \reportchange{\textbf{1.8}} & \newer{\textbf{12.6}} & \newer{\textbf{1.8}} & \newer{\textbf{10.4}} & \newer{\textbf{9.3}} \\
$\Psi_0$ & \newer{2.6} & \reportchange{1.6} & \newer{1.8} & \reportchange{\underline{0.4}} & \newer{1.2} & \reportchange{1.0} & \reportchange{0.6} & \reportchange{1.3} \\
\bottomrule
\end{tabular}
}
\end{table*}
\begin{table*}[!t]
\centering
\small
\caption{
Out-of-distribution generalization on the \emph{Conditional} suite.
We report success rate (\%) under seven controlled OOD settings, averaged over 25 rollouts per task.
\reportchange{Within each column, \textbf{bold} and \underline{underline} mark the two highest distinct positive scores, respectively.}
}
\label{tab:ood_conditional}
\vskip -0.05in
\resizebox{\textwidth}{!}{
\begin{tabular}{lcccccccc}
\toprule
Method & Distractor & Lighting & Background & Camera & Language & Sensor & Robot Init. & OOD Avg. \\
\midrule
Diffusion Policy & \newer{2.4} & \newer{0.0} & \newer{1.6} & \newer{0.4} & \newer{1.6} & \newer{0.4} & \newer{0.0} & \newer{0.9} \\
Flow Matching & \newer{4.8} & \newer{2.4} & \newer{6.8} & \newer{1.2} & \newer{6.0} & \newer{0.4} & \newer{\underline{5.6}} & \newer{3.9} \\
ACT & \newer{8.4} & \newer{0.8} & \newer{5.2} & \newer{0.0} & \newer{5.6} & \newer{0.0} & \newer{3.6} & \newer{3.4} \\
$\pi_{0.5}$ & \newer{\underline{11.2}} & \newer{\underline{12.0}} & \newer{\underline{10.8}} & \newer{\underline{3.2}} & \newer{\underline{10.4}} & \newer{3.6} & \newer{\underline{5.6}} & \newer{\underline{8.1}} \\
GR00T N1.7 & \newer{\textbf{20.8}} & \newer{\textbf{17.6}} & \newer{\textbf{22.0}} & \newer{\textbf{6.8}} & \newer{\textbf{24.8}} & \newer{\textbf{12.4}} & \newer{\textbf{15.2}} & \newer{\textbf{17.1}} \\
$\Psi_0$ & \newer{5.2} & \newer{4.8} & \newer{6.0} & \newer{2.8} & \newer{3.2} & \newer{\underline{4.4}} & \newer{2.8} & \newer{4.2} \\
\bottomrule
\end{tabular}
}
\end{table*}

\subsection{Suite-wise Results of Out-of-Distribution Setting}

As shown in Tables~\ref{tab:ood_basic}--\ref{tab:ood_conditional}, OOD performance varies substantially across HUMANO-Plus suites and distribution shifts. $\pi_{0.5}$ achieves the highest average success rates on the \emph{Basic}, \emph{Advanced}, \emph{Object}, and \emph{Movement} suites, whereas GR00T N1.7 leads on \emph{Spatial}, \emph{Goal}, \emph{Long}, and \emph{Conditional}. Notably, the strongest policies achieve relatively high success rates on \emph{Object} and \emph{Movement}, while performance remains considerably lower on \emph{Goal} and \emph{Long}, highlighting the challenges of goal-conditioned reasoning and long-horizon execution under distribution shifts. Across most suites, \emph{Sensor Noise} and \emph{Camera Viewpoint} introduce particularly severe performance degradation, whereas \emph{Background} and \emph{Language} variations generally yield higher success rates. Moreover, the relative robustness of different policies varies across both suites and OOD settings, indicating complementary rather than uniform generalization capabilities. Overall, these results demonstrate that aggregate OOD performance can obscure substantial differences across capability dimensions and perturbation types, motivating fine-grained robustness evaluation in humanoid loco-manipulation.

\section{Limitations}
\label{app:limitations}

Despite its scale, HUMANO has several limitations. First, the current benchmark primarily focuses on indoor household environments and does not yet cover broader outdoor, industrial, or human-interactive scenarios. Second, although the seven OOD settings capture several important sources of distribution shift, future extensions could further consider variations in physical properties, contact dynamics, terrain conditions, and robot hardware. Third, real-robot evaluation is currently limited to a small number of representative tasks, and broader physical validation remains an important direction for future work.

\end{document}